\documentclass[preprint,12pt]{elsarticle}

\usepackage{amssymb}
\usepackage{amsmath}
\usepackage{multirow}
\usepackage{booktabs}
\usepackage[dvipsnames]{xcolor}
\usepackage{url}

\usepackage{pdflscape}
\usepackage{longtable}   

\journal{Journal of Visual Communication and Image Representation}

\begin{document}

\begin{frontmatter}



\title{Context-Aware Token Pruning and Discriminative Selective Attention for Transformer Tracking}


\author[1,2]{Janani Kugarajeevan}
\ead{jananitha@univ.jfn.ac.lk}

\author[3]{Thanikasalam Kokul\corref{cor1}}
\ead{kokul@univ.jfn.ac.lk}

\author[3]{Amirthalingam Ramanan}
\ead{a.ramanan@univ.jfn.ac.lk}

\author[1]{Subha Fernando}
\ead{subhaf@uom.lk}

\cortext[cor1]{Corresponding author}

\affiliation[1]{organization={Department of Computational Mathematics, University of Moratuwa},
	country={Sri Lanka}}
\affiliation[2]{organization={Department of Interdisciplinary Studies, University of Jaffna},
	country={Sri Lanka}}
\affiliation[3]{organization={Department of Computer Science, University of Jaffna},
	country={Sri Lanka}}

\begin{abstract}
One-stream Transformer-based trackers have demonstrated remarkable performance by concatenating template and search region tokens, thereby enabling joint attention across all tokens.  However, enabling an excessive proportion of background search tokens to attend to the target template tokens weakens the tracker’s discriminative capability. Several token pruning methods have been proposed to mitigate background interference; however, they often remove tokens near the target, leading to the loss of essential contextual information and degraded tracking performance. Moreover, the presence of distractors within the search tokens further reduces the tracker’s ability to accurately identify the target. To address these limitations, we propose CPDATrack, a novel tracking framework designed to suppress interference from background and distractor tokens while enhancing computational efficiency. First, a learnable module is integrated between two designated encoder layers to estimate the probability of each search token being associated with the target. Based on these estimates, less-informative background tokens are pruned from the search region while preserving the contextual cues surrounding the target. To further suppress background interference, a discriminative selective attention mechanism is employed that fully blocks search-to-template attention in the early layers. In the subsequent encoder layers, high-probability target tokens are selectively extracted from a localized region to attend to the template tokens, thereby reducing the influence of background and distractor tokens.  The proposed CPDATrack achieves state-of-the-art performance across multiple benchmarks, particularly on GOT-10k, where it attains an average overlap of 75.1\%. All code, models, and results associated with this work are publicly accessible at \url{https://github.com/JananiKugaa/CPDATrack.git}.
\end{abstract}


\end{frontmatter}

\section{Introduction} \label{sec1}

Visual Object Tracking (VOT) is a fundamental computer vision task that aims to estimate the state of a target object throughout a video sequence, starting from its initial position in the first frame. It is essential for numerous real-world applications, including autonomous driving in dynamic environments \cite{RAHMAN2025101950}, intelligent video surveillance for activity monitoring \cite{ALASIRY2025129281}, mobile robotic operations \cite{REYESREYES2025105311}, augmented reality implementations \cite{Nwobodo2025}, and human-computer interaction systems \cite{King11017161}. Despite significant advancements in tracking approaches over the years, developing a robust and generalizable VOT approach remains challenging due to real-world complexities like occlusions, abrupt motion, illumination changes, cluttered backgrounds, and distractor objects.

Most of the early deep learning-based VOT approaches were Siamese-based convolutional neural network (CNN) trackers \cite{Bertinetto2016,Li_2019_CVPR,zhang2020ocean,voigtlaender2020siam,mayer2021learning}. Although these approaches balance accuracy and efficiency, their reliance on CNNs limits global context modeling, and the use of linear cross-correlation fails to capture complex spatial relationships in challenging scenarios such as occlusion, deformation, and fast motion. To address these limitations, recent trackers have shifted toward Transformer-based models \cite{kugarajeevan2023transformers}, achieving state-of-the-art accuracy through enhanced global context modeling and improved ability to capture long-range dependencies.

Initially, Transformers \cite{vaswani2017attention} were introduced into VOT as a correlation module to enhance feature interaction within CNN-based tracking frameworks. These hybrid CNN–Transformer trackers \cite{chen2021transformer,yu2021high,yan2021learning,song2022transformer,Yang_2023_IEEETrans,ZHANG_2024_CEE,YANG_2024_NN, SUN_2024_KBS,LI_2024_IS,YAO_2024_EAAI}  replace the traditional cross-correlation operation with an attention mechanism, thereby enhancing robustness. However, since they still relied on CNNs for feature extraction, their ability to capture global contextual information is limited. To address this, subsequent trackers evolved into two-stream Transformer architectures \cite{lin2022swintrack,fu2022sparsett,HUANG_2024_EAAI,He_Zhang_Xie_Li_Wang_2023}, eliminating CNNs entirely and processing the template and search regions through separate, identical Transformer encoders. Although two-stream Transformer trackers enable deeper global representation learning, the lack of interaction between features during extraction limits their performance. Recent trackers \cite{ye2022joint, chen2022backbone, cui2022mixformer,Chen2023seqtrack, gao2023generalized,  Yang_2023_ICCV,  Cai_2023_ICCV, Song_Luo_Yu_Chen_Yang_2023, Kang_2023_ICCV, Zhao_2023_CVPR, GONG_2024_KBS,KUGARAJEEVAN2025125381} address this limitation by adopting a single Transformer architecture for simultaneous feature extraction and relation modeling. These one-stream Transformer trackers enhance the tracker’s discriminative ability by jointly processing the template and search patches, thereby achieving state-of-the-art accuracy.

\begin{figure}[t]%
	\centering
	\includegraphics[width=0.48\textwidth]{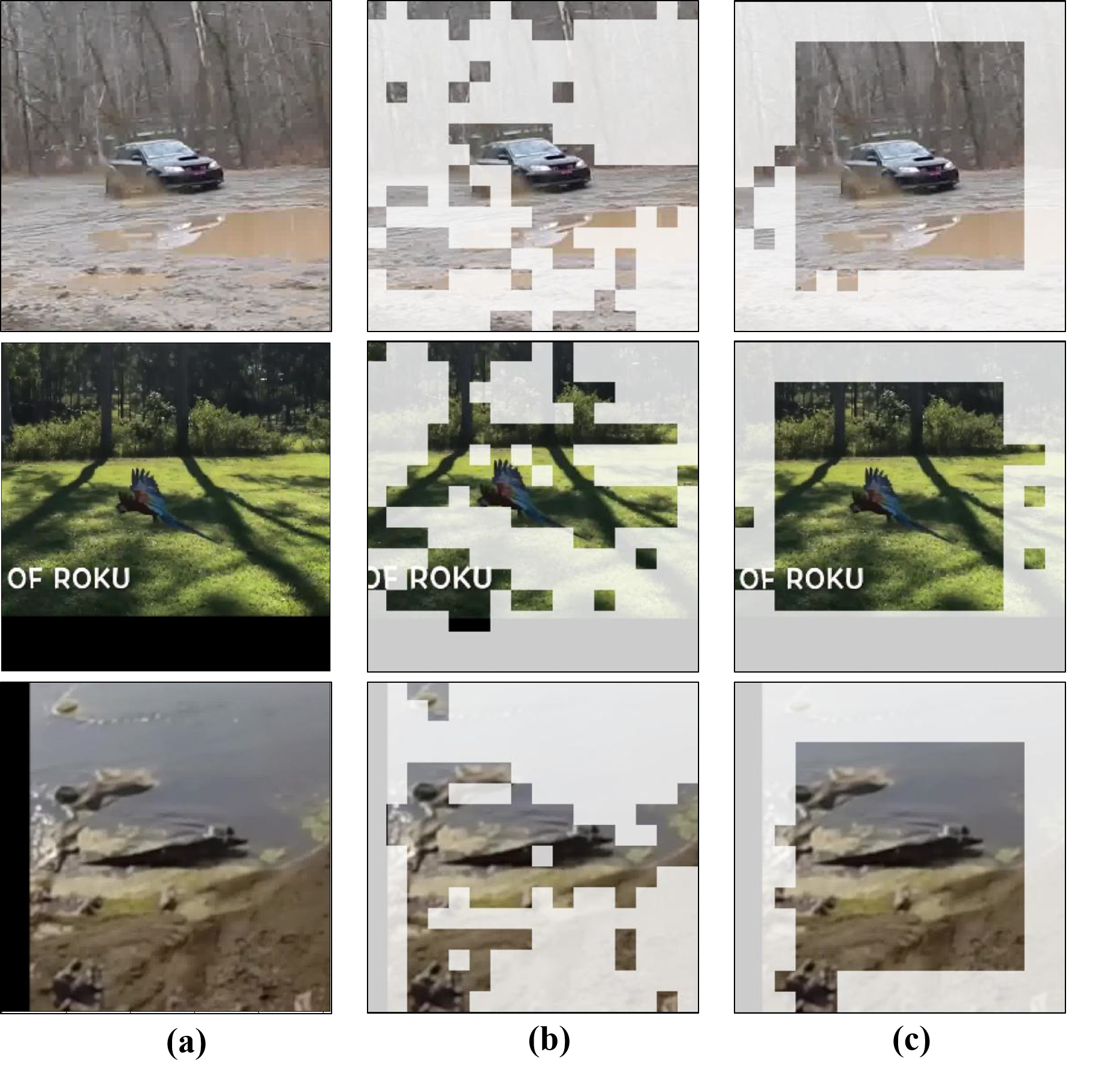}
	\caption{Comparison of token pruning strategies: (a) the actual search region; (b) token elimination in OSTrack \cite{ye2022joint} using conventional background pruning, which may inadvertently discard essential contextual tokens; and (c) the proposed context-aware token pruning method, which preserves a local contextual zone around the target while removing distant background tokens.}
	\label{fig:pruning_comparison}
\end{figure}

One-stream Transformer trackers process the  search region and target template tokens jointly, enabling bidirectional interactions between them through the self-attention mechanism. Although such bidirectional interaction offers several advantages, it also introduces certain challenges. The search region is nearly twice as large as the target template and therefore contains substantial background information, including distractors resembling the target, which contaminate the template features and weaken the model’s discriminative capability. In addition, the large number of search tokens raises the computational burden of the tracker due to the quadratic nature of self-attention with respect to input length, which severely limits the feasibility of real-time tracking.

A few trackers have recognized this issue and proposed token pruning or token fusion strategies to eliminate background tokens from the search region. These token pruning \cite{ye2022joint, Lan10094971, KUGARAJEEVAN2025125381,LIU2025104553} and token fusion methods \cite{XuLiang2024, ZHANG2025105431} identify less informative tokens in the search region based on their attention scores with respect to the center token of the target template, and subsequently eliminate or merge them to improve computational efficiency. Although these trackers significantly enhance tracking efficiency, their accuracy does not improve substantially and, in some cases, even declines due to several factors. First, identifying background tokens solely based on attention scores often generalizes poorly across varying target and background distributions, as these scores may not reliably reflect the true informativeness of the tokens. In addition, as illustrated in Fig. \ref{fig:pruning_comparison}(b), background token elimination methods may also remove nearby target tokens, leading to the loss of essential contextual information required for accurate tracking. Furthermore, fusion-based approaches may inadvertently introduce irrelevant background features, which can blur important spatial details and adversely affect tracking accuracy.

Another group of trackers \cite{gao2023generalized, GONG_2024_KBS, LIU2025112736, KUGARAJEEVAN2025125381} proposed various attention mechanisms to reduce the influence of background tokens in the search region on the target tokens, thereby preserving discriminative target features.  Some trackers \cite{KUGARAJEEVAN2025125381} employed the attention scores of search region tokens relative to the centre token of the target template to identify background tokens, and subsequently suppressed their attention on target tokens in deeper encoder layers to preserve discriminative target feature representations. While this approach enhances accuracy, its sole reliance on attention scores may occasionally result in misclassification of target tokens as background tokens. Another group of trackers \cite{gao2023generalized, GONG_2024_KBS, LIU2025112736} integrated a learnable module that classifies search region tokens into target and background categories at each encoder layer, followed by suppressing the attention of background tokens on the target template. Although this strategy appears promising, the module’s dynamic classification of tokens in each layer often results in the same tokens being identified as target in some layers and as background in others, owing to the varying levels of semantic knowledge across encoder layers. Such inconsistency may disrupt the extraction of target-specific features within the search region. The issue becomes particularly critical in layers where all search tokens are classified as non-target and template-to-search attention is suppressed, thereby reducing tracking accuracy.

In this study, we propose a novel tracking framework to overcome the limitations of existing one-stream Transformer-based trackers. To mitigate background interference in the target template and enhance computational efficiency, a Context-Aware Token Pruning (CATP) module is introduced. As illustrated in Fig.\ref{fig:pruning_comparison}(c), the proposed CATP effectively discards less informative background tokens within the search regions while retaining a local contextual zone surrounding high-probability target regions. The proposed CATP enhances tracking efficiency by discarding background tokens while preserving informative contextual tokens around the target, thereby achieving higher accuracy compared to previous pruning-based approaches. In addition, a Target Probability Estimation (TPE) module is employed in this study to identify high-probability target tokens within the search region and to detect less informative background tokens outside the contextual zone. Unlike previous attention-based token pruning or fusion approaches, the proposed TPE module is a learnable component designed to accurately identify less informative background tokens. It estimates the likelihood of each search token belonging to the target based on its relationship with the max-pooled tokens corresponding to the bounding boxes of the target template and the dynamic template.

To further suppress background interference, we introduce a Discriminative Selective Attention (DSA) mechanism that blocks search-to-template attention in the early encoder layers, as their semantic representations are insufficient to reliably distinguish between target and background tokens. In a middle layer, the TPE module classifies search tokens as either target or background, and this classification remains fixed in the subsequent layers to prevent inconsistency. In each subsequent encoder layer, only a selected set of high-confidence target search tokens is allowed to attend to the template tokens, enabling the tracker to effectively capture dynamic target features from the search region while minimizing background interference on the target template tokens.
Furthermore, to improve the discriminative capability of the tracking model, tokens from distracting objects, even if they have a high target likelihood, are prevented from attending to the target template tokens by selecting target search tokens within a narrow spatial confidence zone centered on the target. 

In summary, our contributions can be summarized as follows:

\begin{enumerate}  
	\item
	A Context-Aware Token Pruning (CATP) mechanism is introduced to discard less informative background tokens within the search region while retaining a local contextual zone around high-probability target regions. Unlike previous approaches, the proposed pruning mechanism is guided by a learnable Target Probability Estimation (TPE) module.  
	
	\item A discriminative selective attention mechanism is designed to suppress background interference by blocking search-to-template attention in the early layers and subsequently allowing only a selected set of target search tokens in the later layers to capture dynamic target features from the search region.

	\item The influence of distractor objects is reduced by selecting target search tokens from a narrow spatial confidence zone centered on the target. As these selected tokens are allowed to attend only to the target template tokens, the discriminative capability of the tracker is significantly enhanced.

	\item Extensive experiments and evaluations are conducted to validate the effectiveness of our tracker. The proposed Context-aware Pruning and Discriminative Attention Tracker (CPDATrack) achieves superior performance across multiple benchmark datasets, including GOT-10k \cite{huang2019got}, LaSOT \cite{Fan_2019_CVPR}, TrackingNet \cite{Muller_2018_ECCV}, and UAV123 \cite{UAV123_2016}.
	
\end{enumerate} 

\section{Related Literature} \label{sec2}
This section reviews the literature on one-stream Transformer-based tracking methods relevant to the proposed approach, with a focus on token pruning, fusion strategies, and attention mechanisms designed to reduce background interference.

\subsection{One-stream Transformer Trackers}\label{sec2sec2}
Over the past four years, one-stream Transformer  trackers \cite{chen2022backbone, cui2022mixformer, Chen2023seqtrack, Yang_2023_ICCV, Cai_2023_ICCV, Song_Luo_Yu_Chen_Yang_2023, Kang_2023_ICCV, Zhao_2023_CVPR} have demonstrated remarkable performance, nearly displacing CNN-based methods in the visual tracking community due to their superior ability to model long-range dependencies. These trackers unify feature extraction and target-search interaction within a single Transformer network, where the self-attention mechanism enables rich bidirectional interactions between template and search tokens, facilitating more adaptive and target-aware feature representations. Most of these trackers employ Vision Transformers \cite{dosovitskiy2021image} as the backbone and are often initialized with models pre-trained using Masked Autoencoders (MAE) \cite{He2022masked}.

Although Transformer-based trackers have achieved state-of-the-art accuracy across various benchmark datasets, several major limitations can still be identified in the literature. Since the large search region typically contains substantial background information, it can dilute the discriminative features of the target template, as both are jointly processed across all encoder layers. Only a few approaches have addressed this issue, while most existing methods have yet to consider it. Furthermore, when similar distractors appear in the search region, they can further weaken the discriminative representation of the target template features, leading to tracking drift. Another major challenge in Transformer-based tracking is its high computational complexity, as the self-attention mechanism scales quadratically with the length of the input sequence. The proposed CPDATrack is designed to address these limitations.

\subsection{Token Pruning and Fusion Approaches}\label{sec2sec2}

To mitigate the interference of background tokens on template tokens and reduce computational complexity, a few one-stream Transformer-based trackers have employed token pruning or token fusion techniques to eliminate less informative background tokens from the search region. 

OSTrack \cite{ye2022joint} was the first tracker to employ token pruning techniques to remove background tokens from the search region. Based on the attention scores between the central target template token and each search region token, OSTrack identifies background tokens and eliminates them by retaining only the top-k tokens with the highest attention scores. This elimination process is applied across multiple layers of the Transformer encoder, specifically during the early, middle, and late stages. However, since OSTrack’s token pruning relies on the attention scores of the initial target template, accurately identifying background tokens in long-term tracking sequences becomes challenging. To address this limitation, ProContext \cite{Lan10094971}, SIFTrack \cite{KUGARAJEEVAN2025125381}, and ATrans \cite{LIU2025104553} incorporated the attention scores of the central dynamic template into the token pruning process. Although the token pruning techniques used in these approaches improve computational efficiency, they provide only limited gains in accuracy, as removing patches around the target in the search region results in the loss of important contextual information. Moreover, background tokens near the target offer crucial spatial context that aids in effective target–background discrimination.

In contrast to token pruning approaches, token fusion trackers merge less informative background tokens to enhance computational efficiency while claiming to preserve contextual information.  Similar to pruning-based approaches, ATFTrack \cite{XuLiang2024} identifies background tokens based on their low attention scores with the central target token and subsequently fuses a significant number of these background tokens together. Similarly, Zhang et al. \cite{ZHANG2025105431} proposed a token fusion approach that classifies all search region tokens into high, medium, and low groups based on their attention scores with the target template.
The low-score background tokens are then eliminated through token pruning, while the medium-score tokens are fused with the high-score tokens. Although these token fusion approaches assert their ability to preserve contextual information, they inadvertently compromise valuable cues, as the fusion of contextually significant background tokens with less informative ones leads to the loss of critical spatial and semantic details. Furthermore, token fusion may still introduce irrelevant background features, which can obscure crucial spatial details and degrade tracking accuracy.

In both token pruning and fusion-based approaches, background tokens are typically identified based on their low attention scores with central target or dynamic template tokens. However, attention scores are highly dependent on the underlying feature distribution, positional biases, and layer depth, and therefore do not necessarily correspond to semantic relevance.  In contrast to these methods, the proposed CPDATrack prunes background tokens under the guidance of a learnable Target Probability Estimation (TPE) module. This learnable module adapts to variations in target appearance, enabling a more robust and generalisable decision boundary for distinguishing target and background tokens. Furthermore, by preserving tokens in the vicinity of the target, the proposed tracker maintains essential contextual information.

\begin{figure*}[t!]
	\centering
	\includegraphics[width=0.98\textwidth]{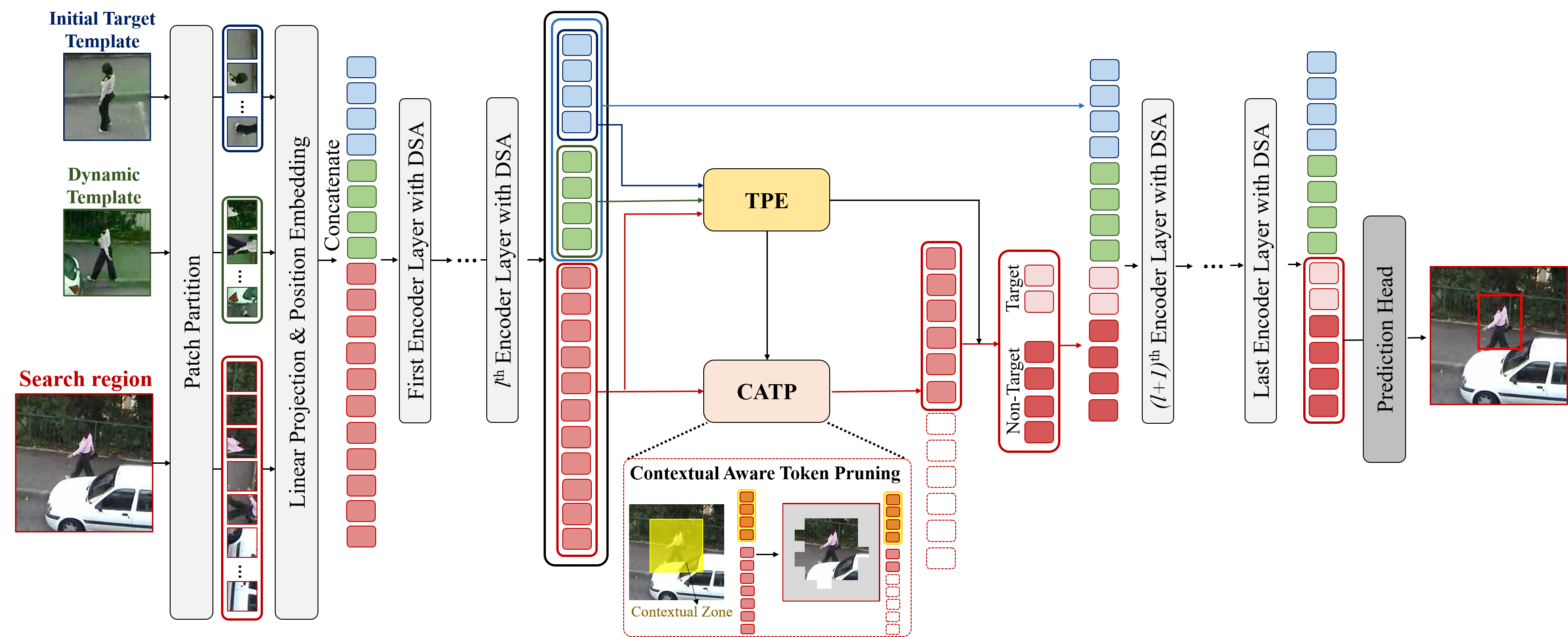}
	\caption{The overall architecture of the proposed CPDATrack approach. The Target Probability Estimation (TPE) module predicts the target likelihood of each search token based on the target template and dynamic template tokens. Guided by these predictions, the Context-Aware Token Pruning (CATP) module removes less informative background tokens in the search region while preserving essential contextual information. The Discriminative Selective Attention (DSA) Mechanism  further reduces the interference of background tokens on the target and dynamic template tokens.}
	\label{fig:Methodology}
\end{figure*}

\subsection{Background Interference Reduction Approaches}\label{sec2sec2}

One-stream Transformer trackers jointly process both the search region and target template tokens within each encoder layer. Although bidirectional attention between the template and search region offers several advantages, interference from the large number of background tokens in the search region can degrade the target representation and significantly weaken the model’s discriminative capability. A few trackers have recognised this issue and proposed various techniques to mitigate background interference without relying on token pruning or fusion.

GRM \cite{gao2023generalized}, ASAFormer \cite{GONG_2024_KBS}, and RIDTrack \cite{LIU2025112736} employed learnable modules to categorise search region tokens into target and non-target groups within each encoder layer. Subsequently, only the target search tokens were allowed to perform cross-attention with the template tokens, thereby reducing interference from background tokens. Although this module improves performance, the semantic representations of encoder layers differ from one another; as a result, a token may be identified as a target in one layer but as background in another. Such inconsistent classification can disrupt feature learning and may lead to degraded tracking performance.

SIFTrack \cite{KUGARAJEEVAN2025125381} entirely blocks search-to-template attention in the early encoder layers to prevent background interference, and in the deeper layers, only the top-ranked target search tokens are allowed to perform cross-attention. Although this tracker maintains the same set of target and non-target token groups across all remaining layers, its reliance on attention score–based classification rather than a learnable module introduces inherent limitations. Furthermore, SIFTrack selects a fixed number of target search tokens across all tracking sequences, which may lead to the inclusion of irrelevant background tokens when the target is small or heavily occluded, while potentially omitting important tokens for larger targets.

Recent trackers, such as F-BDMTrack \citep{Yang_2023_ICCV}, ROMTrack \citep{Cai_2023_ICCV}, and ASTrack \citep{CHEN2025126910}, have also proposed a range of mechanisms aimed at mitigating the impact of background interference. 
F-BDMTrack  employs a foreground–background agent learning module to generate distinct representations for the template and search regions and suppresses incorrect interactions between the foreground and background by modeling the distribution similarity of each patch feature with the learned foreground–background representations through a distribution-aware attention mechanism. ROMTrack  mitigates background interference on the target template by completely blocking search-to-target template cross-attention across all layers, while permitting only search-to-dynamic template cross-attention. ASTrack   reduces background interference by generating auxiliary search tokens based on their similarity to the template and introducing a bidirectional interactive modulation module, where the interaction between general and auxiliary tokens aggregates target-relevant features and suppresses background interference. Although these trackers considerably mitigate background interference, certain limitations remain. F-BDMTrack employs a pseudo-bounding-box generation strategy to distinguish foreground and background tokens within the search region; however, this approach may yield inaccurate pseudo-labels and introduce noise into the learning process. In contrast, both the dynamic template in ROMTrack and the target template in ASTrack are jointly processed with the search region, which allows residual background interference and consequently degrades the fidelity of the target representation.

All previous approaches have employed either a learnable module or an attention-based mechanism to identify potential target tokens within the search region and subsequently mitigate background interference by restricting their interaction to template tokens. However, distractor objects within the search region may still be erroneously identified as target tokens due to their feature similarity with the template, thereby diluting the discriminative features of the true target. To date, no prior studies have explicitly addressed this limitation. 

In this work, a novel discriminative selective attention mechanism is proposed to mitigate the influence of background tokens on target template tokens. Furthermore, by restricting cross-attention to a compact spatial region surrounding high-confidence target tokens within the search area, the influence of distractor objects is mitigated, and the discriminative features of the target template are effectively preserved.

\section{Methodology} \label{sec3}

In this work, a novel one-stream tracking framework is proposed, termed CPDATrack. This section provides a detailed description of the proposed tracker and its key modules.

\subsection{Overview} \label{sec3sec1}

The overall architecture of the proposed CPDATrack is illustrated in Fig. \ref{fig:Methodology}. It takes three inputs: an initial target template image $ZI \in \mathbb{R}^{ H_{ZI} \times W_{ZI} \times 3}$, a dynamic template image $ZD \in \mathbb{R}^{H_{ZD} \times W_{ZD} \times 3}$ and a search region image $X \in \mathbb{R}^{H_X \times W_X \times 3}$ , where $ H_{ZI} \times W_{ZI} $, $ H_{ZD} \times W_{ZD} $ and  $H_X \times W_X$ denotes the spatial dimensions of the initial target template, dynamic template and search image, respectively. Similar to other Transformer-based trackers, the inputs are first divided into non-overlapping patches of size $P \times P$. These patches are then projected into patch embeddings, and learnable positional embeddings are added to form the inital template tokens $E_{ZI} \in \mathbb{R}^{N_{ZI} \times D} $ , dynamic template tokens $E_{ZD} \in \mathbb{R}^{N_{ZD} \times D} $and search tokens $E_X \in \mathbb{R}^{N_X \times D}$.  Here, $N_{ZI} = \frac{H_{ZI} \times W_{ZI} }{P^2}$, $N_{ZD} = \frac{H_{ZD} \times W_{ZD} }{P^2} $ and $N_X = \frac{H_X \times W_X }{P^2}$ denote the number of tokens in each input, and $D$ is the token dimension. All input tokens are then concatenated to form a combined token sequence [$E_{ZI}$;$E_{ZD}$;$E_X$] with a total length of $N_{ZI}+N_{ZD}+N_X$. The concatenated tokens are subsequently processed by a Transformer composed of $L$ stacked encoder blocks for feature extraction. 

As shown in Fig. \ref{fig:Methodology}, the proposed tracker comprises several key modules designed to enhance its discriminative capability and computational efficiency. In the early encoder layers (from the first layer up to layer $l$), a Discriminative Selective Attention (DSA) mechanism is employed to mitigate the interference of background tokens in the search regions on both the initial and dynamic target templates. Between two particular early encoder layers, denoted as $l^{th}$ and  $(l+1)^{th}$ layers in Fig. \ref{fig:Methodology}, a Target Probability Estimation (TPE) module is incorporated to estimate the target likelihood of each search token based on the features of the target and dynamic templates. In addition, a Context-Aware Token Pruning (CATP) module is also included between these two layers, which removes less informative background tokens in the search region while preserving essential contextual information under the guidance of the TPE module. After background token pruning, the remaining search region tokens are classified as target or non-target tokens under the guidance of the TPE module. In the remaining encoder layers, the DSA mechanism is re-employed, treating the remaining tokens as initial and dynamic target template tokens, target search tokens, and non-target search tokens.

Finally, after the last encoder layer, the search region tokens are passed to the prediction head for target localisation. The details of each module and mechanism are described in the following subsections.

\subsection{Target Probability Estimation Module} \label{sec3sec2}

\begin{figure}[t]%
	\centering
	\includegraphics[width=0.48\textwidth]{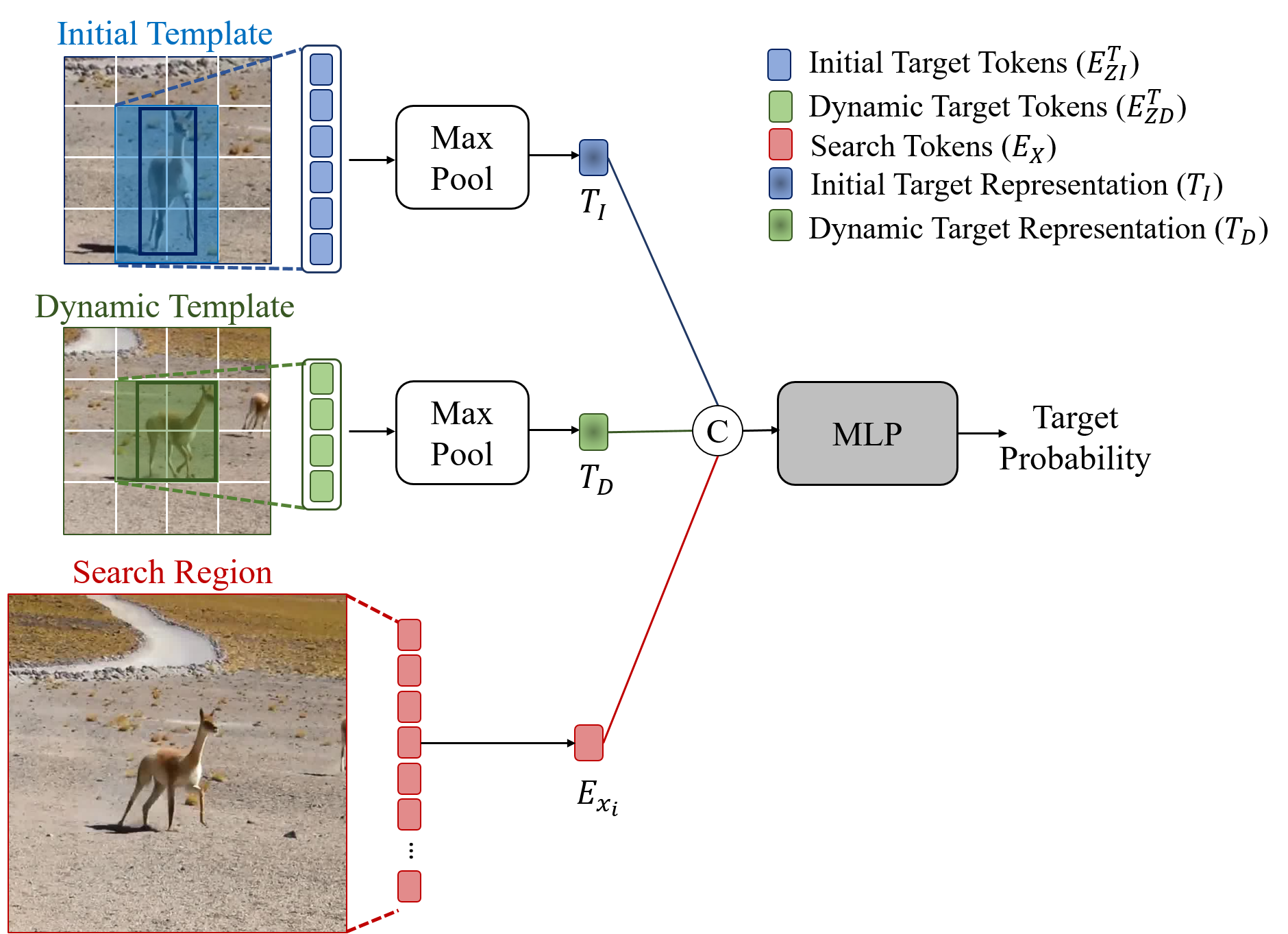}
	\caption{Proposed Target Probability Estimation (TPE) module. True target representations from the initial and dynamic templates are concatenated with each search token to predict the likelihood of it belonging to the target.}
	\label{fig:TargetProbabilityEstimation}
\end{figure}

The proposed Target Probability Estimation (TPE) module is designed to estimate the likelihood of the target’s presence within each search region token. It plays a vital role in the proposed tracker by guiding both the pruning module and the discriminative selective attention mechanism. As illustrated in Fig.\ref{fig:Methodology}, the TPE module is integrated after the initial few encoder layers, as the early-layer representations are insufficient for reliably estimating the target’s presence across search region tokens. The specific placement of the TPE module between encoder layers is determined based on empirical analysis. 

The proposed TPE module estimates the probability of the target’s presence in each search region token by learning the relationships among the search region token, the initial target template tokens, and the dynamic target template tokens. Unlike previous approaches \cite{gao2023generalized, LIU2025112736, GONG_2024_KBS}, the dynamic target template tokens are also incorporated into the probability estimation process, as they play a crucial role in capturing variations in the target’s appearance. First, the features of the target template, dynamic template, and search region are obtained from the encoder layer $l$. However, as the initial target and dynamic templates also contain a considerable number of background tokens, as illustrated in Fig. \ref{fig:TargetProbabilityEstimation}, only the tokens corresponding to the actual target (denoted as $E_{ZI}^{T}$ and $E_{ZD}^{T}$) are selected based on their bounding box coordinates. This selection process ensures that the TPE module learns a true target representation without background interference. These selected initial and dynamic template tokens are then individually aggregated using global maximum pooling to produce compact and informative target representations (denoted as $T_I$ and $T_D$) for both the initial and dynamic templates. This entire process can be written as:

\begin{equation}
	\label{eqn:TargetRepresentation}
	\begin{split}
		& T_I = \text{MaxPool}(E_{ZI}^{T}), \quad E_{ZI}^{T} = \mathcal{B}(E_{ZI})  \\
		& T_D = \text{MaxPool}(E_{ZD}^{T}), \quad E_{ZD}^{T} = \mathcal{B}(E_{ZD}) 
	\end{split}
\end{equation}

\noindent
where $\mathcal{B}(\cdot)$ denotes the operation that selects the subset of template tokens located within the target’s bounding box region.

The max-pooled initial target and dynamic target features, $T_I$ and $T_D$, represent the essential characteristics required to identify the target. Subsequently, the initial and dynamic target representations are concatenated with each search token $E_{x_i}$ and fed into a lightweight multi-layer perceptron (MLP) network. This MLP is designed to predict the likelihood of the target’s presence ($p_i$) within each search token $E_{x_i}$ as follows:

\begin{equation}
	\label{eqn:MLP}
	p_i = \sigma(MLP([T_I;T_D;E_{x_i}])) \quad \forall i \in \{1, \dots, N_X\}
\end{equation}

where $\sigma(.)$ is the sigmoid activation function. This probability estimation is subsequently utilised to guide both the pruning module and the discriminative selective attention mechanism.

\begin{figure*}[t]%
	\centering
	\includegraphics[width=0.98\textwidth]{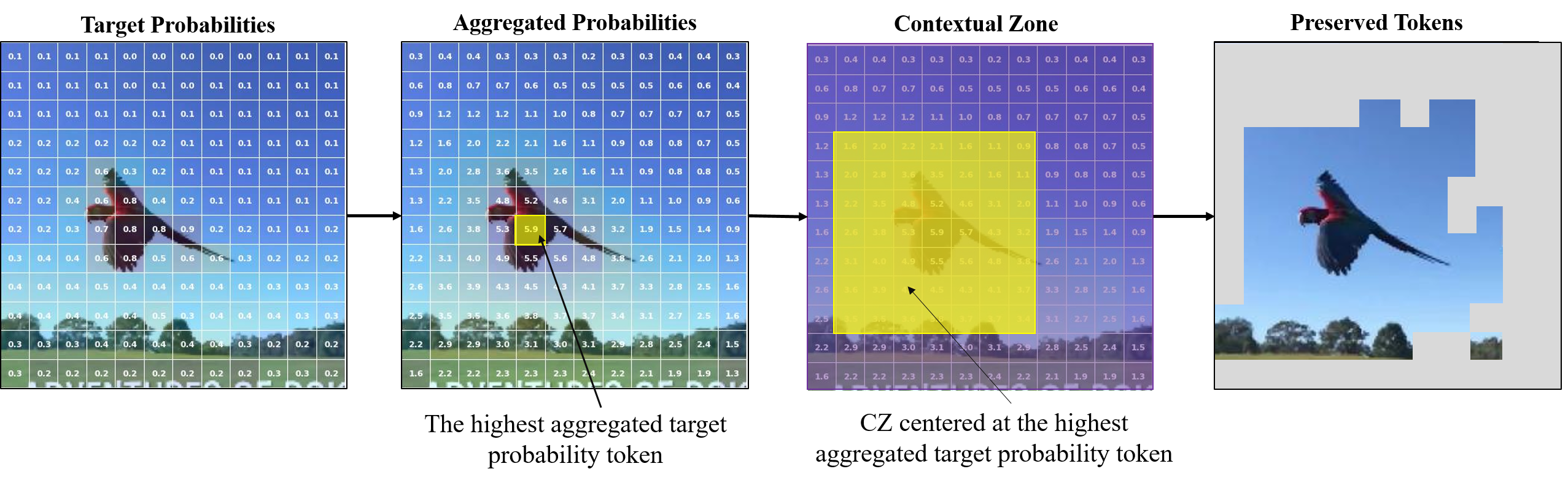}
	\caption{The proposed Context-Aware Token Pruning (CATP) module. The target probabilities of the search region tokens are first aggregated using a sliding $3 \times 3$ window. Subsequently, a contextual zone (CZ) is defined, centered on the token with the highest aggregated target probability, and a fixed number of background tokens outside the CZ are pruned.}
	\label{fig:Contextual_aware_pruning}
\end{figure*}

\subsection{Context-Aware Token Pruning Module} \label{sec3sec3}

In Transformer-based tracking, each token in the search region is treated as a potential target candidate. However, a considerable proportion of these tokens originate from background regions, introducing noise during feature learning and increasing computational overhead. Although previous studies have attempted to mitigate this limitation through token pruning, excessive pruning often results in the loss of essential contextual information surrounding the target, thereby reducing tracking accuracy. To address this limitation, the proposed tracker introduces a Context-Aware Token Pruning (CATP) module. 

The proposed CATP module prunes less informative background tokens while retaining those surrounding the target, as they provide essential contextual cues to the tracker, as illustrated in Fig. \ref{fig:Contextual_aware_pruning}. Since the CATP module preserves essential contextual information while discarding distant background tokens, it enhances the tracker’s discriminative capability and substantially reduces computational complexity. 

The CATP module is integrated into the proposed tracker alongside the TPE module and is positioned between the encoder layers $l$ and $l+1$. Initially, the CATP module receives the likelihood estimates of the target’s presence for all search region tokens from the TPE module. Based on these estimates, a 2D spatial probability map
$P \in \mathbb{R}^{\sqrt{N_X} \times \sqrt{N_X}}$
is generated, where each element $P(u,v)$ represents the target probability of the token located at spatial position $(u,v)$ within the search region. 

In the subsequent stage of the CATP module, the token with the highest target probability within the search region is identified. To accomplish this, a $3 \times 3$ sliding window is applied over the spatial probability map $P$. At each spatial location, the aggregated probability is calculated by summing the probabilities of all tokens within the window, with zero-padding applied to accommodate boundary regions. The aggregated sum ($S$) at a window centered at $(u,v)$ is defined as:

\begin{equation}
	S_{(u,v)} = \sum_{i=-1}^1 \sum_{j=-1}^1 \mathbf{P}_{(u+i,v+j)},
\end{equation}

This aggregation process captures local contextual information and mitigates the influence of isolated tokens with spuriously high probabilities. The token corresponding to the window centered at $(u^*,v^*)$, which exhibits the maximum aggregated probability, is selected as the highest target probability token in the search region.

In the final stage of the CATP module, a contextual zone (CZ) comprising $n \times n$ tokens is defined, centered at the highest aggregated target probability token $(u^*,v^*)$ in the search region. Since the CZ encompasses both high-probability target tokens and background tokens containing valuable contextual information, it is preserved from pruning. Among the tokens outside the CZ, $t$ number of tokens with the lowest target probabilities are identified and subsequently pruned from the search region. In this pruning process, $n$ and $t$ are hyperparameters determined experimentally.

Since the proposed CATP module identifies and prunes a considerable number of less informative background tokens in an early middle encoder layer, the subsequent encoder layers process fewer search region tokens, thereby improving the computational efficiency of the tracker. Furthermore, by reducing the interference of background search tokens on target search tokens, the module enhances tracking accuracy and enables the tracker to more precisely localize the target in the final encoder layer.

\subsection{Discriminative Selective Attention Mechanism} \label{sec3sec4}
\begin{figure}[t]%
	\centering
	\includegraphics[width=0.48\textwidth]{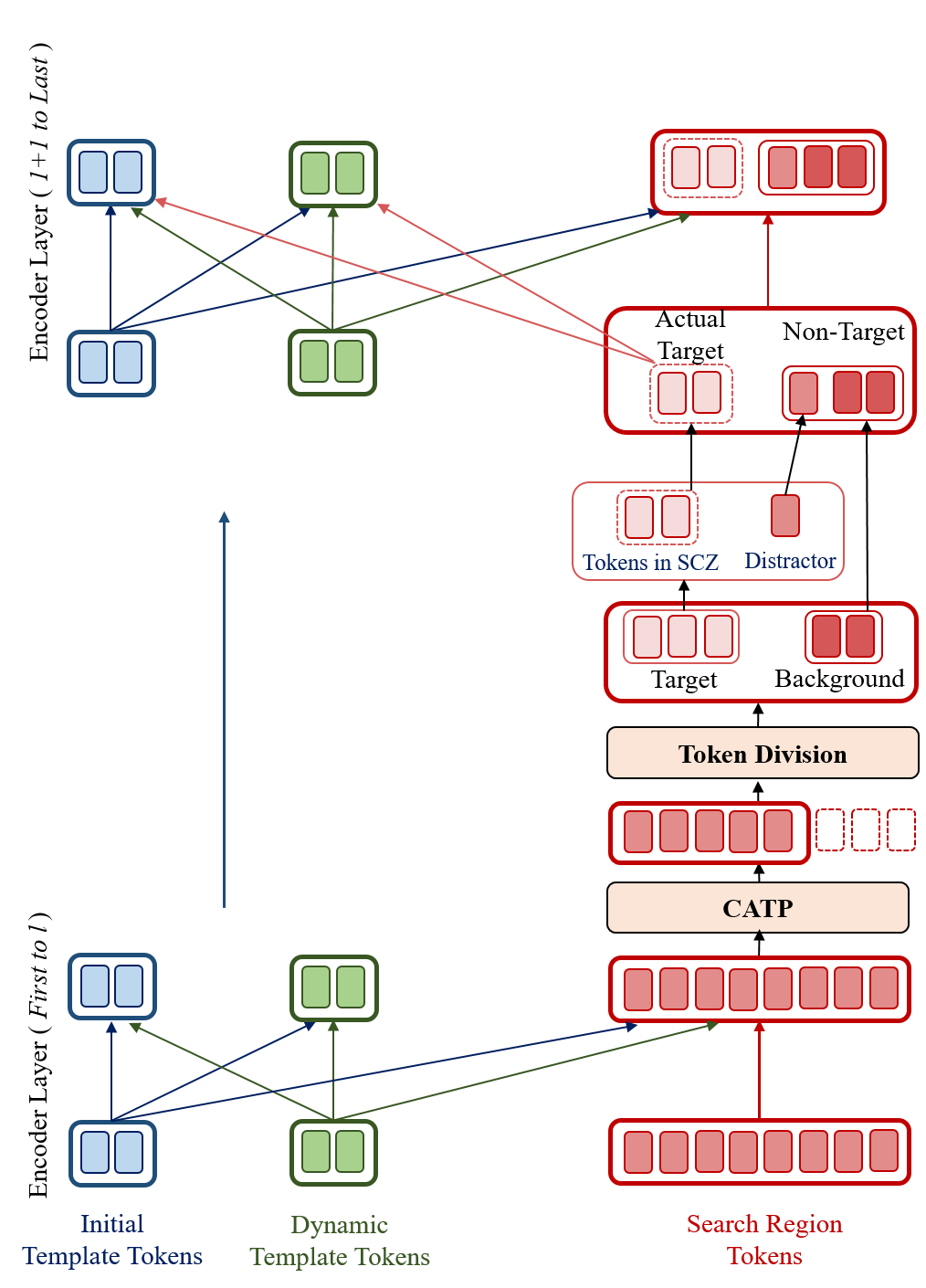}
	\caption{Overview of the proposed Discriminative Selective Attention (DSA) mechanism. In the early encoder layers, search-to-template cross-attention is suppressed to prevent background interference. After token pruning, search tokens are first divided into target and background tokens. Subsequently, target tokens are further classified into actual target tokens and distractor tokens. In the remaining layers, only the actual target tokens are permitted to perform cross-attention with both the initial and dynamic template tokens.}
	\label{fig:SAM}
\end{figure}

In one-stream Transformer tracking, tokens from the initial target template, dynamic target template, and search region are concatenated and jointly processed through a self-attention mechanism. Since all tokens are treated equally and attend to each other across all encoder layers, the large number of background tokens from the search region tends to weaken the discriminative features of the initial and dynamic target template tokens, thereby reducing the tracker’s performance. Furthermore, the presence of distractor object tokens within the search region can also interfere with the target template tokens, further diminishing their discriminative representation.  Although a few existing approaches have recognized these issues, they have not effectively addressed them. To overcome these limitations, the proposed tracker incorporates a Discriminative Selective Attention (DSA) mechanism, which is integrated into all encoder layers of the proposed tracker.

\begin{figure}[t]%
	\centering
	\includegraphics[width=0.48\textwidth]{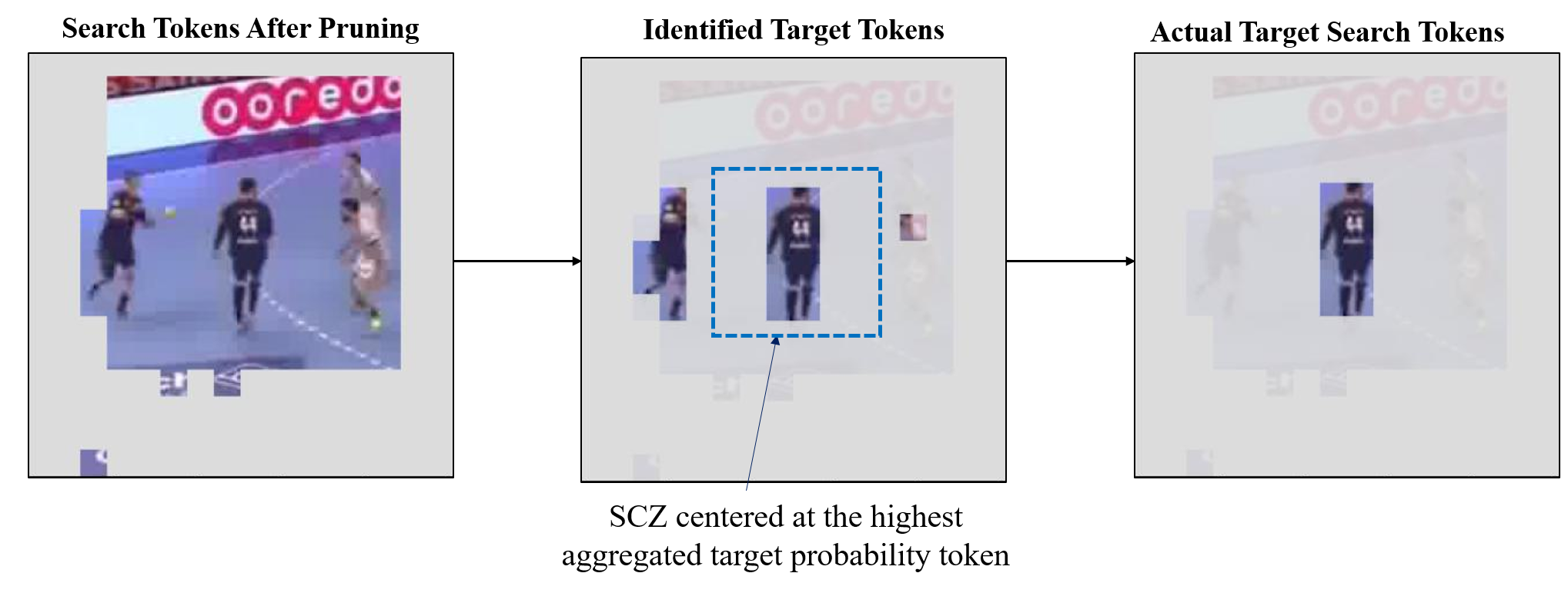}
	\caption{Illustration of how the proposed DSA mechanism mitigates the influence of distractors. After pruning the background tokens, the TPE module identifies target search tokens. However, as shown in the middle image, certain distractors may still be misclassified as targets due to visual similarities. To address this, a Spatial Confidence Zone (SCZ) is established around the high-confidence target token, and only the actual target tokens within the SCZ are permitted to attend to the initial target and dynamic target template tokens.}
	\label{fig:SCZ}
\end{figure}

The proposed CPDATrack framework processes the initial template tokens ($E_{ZI}$), dynamic template tokens ($E_{ZD}$), and search region tokens ($E_{X}$). As illustrated in Fig. \ref{fig:SAM}, in the early layers of the tracker, up to the encoder layer where the TPE and CATP modules are integrated, the proposed DSA mechanism blocks the attention of all search region tokens toward the initial and dynamic templates. The cross-attention mechanism employed in the initial encoder layers can be expressed as:

\begin{equation}
	\label{eq:dsa_early}
	\begin{split}
		& E_{ZI},\, E_{ZD} \rightarrow E_{X} \\
		& E_{ZI} \leftrightarrow E_{ZD} \\
	\end{split}
\end{equation}

where $\rightarrow$ denotes the  one-way cross-attention, and  $\leftrightarrow$ represents the mutual cross-attention. This search to template cross-attention blocking prevents interference from the large number of background tokens in the search region. Although the attention of target search tokens to the initial and dynamic template tokens is important for learning target appearance variations, it is also suppressed at this stage since the early encoder layers do not have sufficient semantic knowledge to effectively distinguish target tokens from background tokens in the search region.

During the integration of the TPE and CATP modules between layers $l$ and $l+1$, the TPE module first estimates the probability of the target’s presence within each search region token. Subsequently, the CATP module prunes a substantial number of background tokens based on these probability estimates. Although the interference from background tokens is considerably reduced through pruning, the DSA is integrated into the remaining encoder layers to further mitigate such interference and minimize the influence of distractor objects.  

Before being fed into the $(l+1)$ encoder layer, the remaining search region tokens are classified as target or background tokens based on their probability scores generated by the TPE module. Since the target search tokens ($E_{XT}$) contain informative, target-specific cues, they should be permitted to attend to both the initial template and dynamic template tokens, whereas the background search tokens ($E_{XB}$) should be restricted to minimise background interference. However, as illustrated in Fig. \ref{fig:SCZ}, certain distractor tokens may be incorrectly classified as target tokens by the TPE module due to their feature similarity to the actual target. Consequently, allowing these distractor tokens to attend to the initial and dynamic target templates can diminish the discriminative capability of the learned feature representations. 

To reduce the influence of distractor tokens, the target search tokens ($E_{XT}$) are further categorised into actual target tokens ($E_{XAT}$) and distractor tokens ($E_{XD}$). This categorisation is achieved by defining a narrow Spatial Confidence Zone (SCZ) of window size $m \times m$, centred on the token with the highest aggregated target probability. Target search tokens located outside the SCZ are thus regarded as distractors. Here, $m$ is a hyperparameter determined empirically.

In the remaining encoder layers, from layer $l+1$ to the final layer, the DSA mechanism is applied by considering the tokens into four groups: initial  template tokens ($E_{XI}$), dynamic  template tokens ($E_{XD}$), actual target search tokens ($E_{XAT}$), and non-target search tokens ($E_{XNT}$). The non-target search tokens comprise both background search tokens ($E_{XB}$) and distractor tokens ($E_{XD}$). In the remaining layers, the attention of all non-target search tokens towards the initial template and dynamic template tokens is blocked to minimise interference from background and distractor objects. The proposed DSA mechanism in these layers can be formulated as: 

\begin{equation}
	\label{eq:dsa_later}
	\begin{split}
		& E_{ZI},\, E_{ZD} \rightarrow E_{XNT} \\
		& E_{ZI},\, E_{ZD} \leftrightarrow E_{XAT} \\
		& E_{ZI} \leftrightarrow E_{ZD}
	\end{split}
\end{equation}

The proposed DSA mechanism effectively reduces the interference of background and distractor tokens on the initial and dynamic target template tokens, thereby enhancing the discriminative capability of the tracker.  

\subsection{Prediction head}
Following the final encoder layer, the search region features are reshaped into a two-dimensional spatial map of size $H_X \times W_X$ and fed into the prediction head to localize the target. Since most background tokens are pruned in the early middle layers, the original spatial arrangement of tokens becomes disrupted, making it infeasible to directly reshape the refined tokens into a 2D feature map. As the discarded tokens correspond to irrelevant background regions and have no impact on the final prediction, they are replaced with zero-padded placeholders before being fed into the prediction head. 

Similar to most one-stream Transformer-based trackers, the prediction head of the proposed tracker comprises three modules: a classification head that outputs the target score map, an offset head that predicts local offsets to correct discretization errors, and a size estimation head that predicts the normalized bounding box dimensions. The total loss is a weighted combination of the individual losses, including a focal loss \cite{law2018cornernet} for the classification head ($L_{cls}$), a generalized IoU loss ($L_{giou}$) \cite{rezatofighi2019generalized} for the offset head, and an $L_1$ loss for the size head. The total loss is defined as: 
\begin{equation}
	\label{eqn:loss}
	Loss = 	L_{cls} +\lambda_{iou}L_{iou} + \lambda_{L_1}L_{1}
\end{equation}

where the regularization coefficients are set to $\lambda_{iou} = 2$ and $\lambda_{L_1} = 5$.

\section{Experiments}\label{sec4}

\subsection{Implementation Details}\label{sec4subsec1}
The proposed tracking framework is based on a ViT-B \cite{dosovitskiy2021image} backbone, which is initialized using pretrained parameters from the MAE \cite{He2022masked} model. The input size is set to $128 \times 128$ for both the target and dynamic templates, and $256 \times 256$ for the search region. Despite other trackers employing both a small $256 \times 256$ and a larger $384 \times 384$ search region, we restrict our experiments to the smaller variant due to GPU memory limitations.

The proposed TPE and CATP modules are integrated between the fourth and fifth encoder layers. In the CATP module, a contextual zone (CZ) with a window size of $11 \times 11$ is defined, centered on the token with the highest aggregated target probability.  The CATP module prunes 128 tokens in every frame of the tracking sequence. In the DSA module, the Spatial Confidence Zone (SCZ) is centered on the token with the highest aggregated target probability, with a window size of $7 \times 7$.

The proposed CPDATrack is trained on large-scale tracking datasets, including GOT-10k \cite{huang2019got}, TrackingNet \cite{Muller_2018_ECCV},  and COCO 2017 \cite{coco_2014}, LaSOT \cite{Fan_2019_CVPR}, and evaluated its performance on TrackingNet, UAV123, and LaSOT test sets. Following the GOT-10k protocol \cite{huang2019got}, we train the model using only the GOT-10k training split and evaluate its performance on the corresponding test set. The proposed tracker is trained for 400 epochs, with each epoch containing 60k image pairs, using the AdamW \cite{loshchilov2018decoupled} optimizer. The learning rate starts at 1e-4 and is reduced by a factor of 10 after the 240th epochs. For training on GOT-10k, the model is trained for 100 epochs, and the learning rate is decayed after 80 epochs.

The proposed tracker is implemented in the PyTorch framework, and all training and evaluation are performed on a Tesla P100 GPU equipped with 16 GB of memory.

\subsection{Evaluation Metrics}\label{sec4subsec2}
To evaluate the performance of the proposed tracker, the standard metrics used in visual object tracking benchmarks were followed. or GOT-10k, the average overlap (AO), success rate at a threshold of 0.5 (SR$_{0.5}$), and success rate at a threshold of 0.75 (SR$_{0.75}$) are used in accordance with the dataset’s evaluation protocol. For LaSOT and TrackingNet, we adopt the area under the curve (AUC), precision (P), and normalized precision (P$_n$). On UAV123, tracking performance is evaluated using AUC and precision (P).

\subsection{Experimental Results and Comparisons}

The tracking performance of the proposed CPDATrack was comprehensively assessed on four widely recognised benchmark datasets: GOT-10k \cite{huang2019got}, TrackingNet \cite{Muller_2018_ECCV}, UAV123 \cite{UAV123_2016}, and LaSOT \cite{Fan_2019_CVPR}. The performance of CPDATrack was compared with state-of-the-art one-stream Transformer trackers \cite{ye2022joint,chen2022backbone,cui2022mixformer,Chen2023seqtrack,gao2023generalized,Yang_2023_ICCV,Cai_2023_ICCV,Song_Luo_Yu_Chen_Yang_2023,Kang_2023_ICCV,Zhao_2023_CVPR,GONG_2024_KBS,KUGARAJEEVAN2025125381,XuLiang2024,ZHANG2025105431,ZHANG2025113230,CHEN2025126910,CHENG2025128821,Zhu10947615,ZHANG2025125654,LIU2025104553,WU2025104608}, two-stream Transformer trackers \cite{HUANG_2024_EAAI,He_Zhang_Xie_Li_Wang_2023,WANG2025129417,Zhu_Tang_Chen_Wang_Wang_Lu_2025}, CNN-Transformer trackers\cite{chen2021transformer,yu2021high,yan2021learning,Yang_2023_IEEETrans,ZHANG_2024_CEE,YANG_2024_NN,SUN_2024_KBS,YAO_2024_EAAI,SUN2025128785,ZHAO2025111,Gopal_2024_WACV,Liang_Li_Long_2023,wang2021transformer,YU2024104067}, and a CNN tracker  \cite{ZHANG_2024_PR,YANG2024104237}  on the GOT-10k, TrackingNet, and UAV123 datasets, and the results are presented in Table \ref{MasterTable}.

GOT-10k is a well-known dataset in visual object tracking (VOT), characterized by its non-overlapping training and testing object classes and a large diversity of object categories. Moreover, since the evaluation protocol restricts training to the GOT-10k train set to prevent familiar class bias, the experimental results on GOT-10k are particularly significant. The performance of our tracker is reported using results from the dataset’s evaluation server. As shown in Table \ref{MasterTable}, the proposed CPDATrack achieves superior performance on the GOT-10k dataset, outperforming recent state-of-the-art trackers. In particular, the proposed tracker outperforms other token pruning–based trackers, including OSTrack-256 \cite{ye2022joint}, ATFTrans \cite{XuLiang2024}, PTFTrack \cite{ZHANG2025105431}, and SIFTrack \cite{KUGARAJEEVAN2025125381}. It also surpasses previous studies that introduced various techniques to mitigate background interference, such as GRM \cite{gao2023generalized}, ASAFormer \cite{GONG_2024_KBS}, F-BDMTrack-256 \cite{Yang_2023_ICCV}, ROMTrack \cite{Cai_2023_ICCV}, and ASTrack \cite{CHEN2025126910}. The experimental results demonstrate the effectiveness of the proposed context-aware token pruning and discriminative selective attention mechanisms.

TrackingNet includes a large collection of tracking sequences featuring numerous target object classes, varied motion dynamics, and complex real-world conditions, providing a comprehensive benchmark for assessing tracker performance and robustness. As shown in the experimental results table \ref{MasterTable}, our tracker outperformed state-of-the-art approaches, attaining a peak AUC of 84.1\%, the second-best precision score of 88.9\%, and the third-top precision score of 82.8\%. The outstanding performance of proposed CPDATrack on the TrackingNet dataset shows its strength in real-world tracking scenarios, effectively handling diverse object classes, rapid motion, scale variations, and occlusions.

The UAV123 dataset consists of aerial tracking sequences that involve monitoring objects from a UAV perspective. The targets are typically small and frequently experience rapid changes in position, scale, and viewpoint due to UAV motion. Additional challenges include partial occlusions, camera vibrations, and complex backgrounds. On the UAV123 dataset, the proposed tracker achieved strong performance, obtained an AUC of 69.4\% and the second-highest precision of 90.6\%, comparable to state-of-the-art approaches. These results highlight its robustness and accuracy in handling challenging sequences with small target objects.

The LaSOT dataset comprises 280 long-term tracking sequences, each with an average length of approximately 2,500 frames, covering a diverse range of object categories and real-world scenarios. As reported in Table \ref{Lasot}, proposed CPDATrack achieved strong performance on LaSOT, with a top normalized precision score of 78.8\%, the third-highest AUC score of 69.1\%, and a precision score of 74.9\%. Overall, these findings underscore the effectiveness of CPDATrack in handling long-term tracking challenges, confirming its ability to generalize across diverse scenarios while maintaining robust and precise performance throughout extended sequences.

\begin{landscape}
	\footnotesize
	\setlength{\LTcapwidth}{\linewidth}
\begin{longtable}{lcl@{\extracolsep{0.30cm}}l@{\extracolsep{0.30cm}}l@{\extracolsep{0.4cm}}l@{\extracolsep{0.30cm}}l@{\extracolsep{0.30cm}}l@{\extracolsep{0.4cm}}l@{\extracolsep{0.30cm}}l@{\extracolsep{0.30cm}}}
	
	\caption{Comparison of proposed CPDATrack with state-of-the-art trackers on the GOT-10k, TrackingNet, and UAV123 datasets using the evaluation metrics $\mathrm{AO}$ (Average Overlap), $\mathrm{SR_{0.5}}$ and $\mathrm{SR_{0.75}}$ (Success Rates at 0.5 and 0.75 thresholds), $\mathrm{AUC}$ (Area Under the Curve), $\mathrm{P}$ (Precision), and $\mathrm{P_n}$ (Normalized Precision). All results are reported as percentages. Since the proposed tracker uses a search region size of $256 \times 256$, the same setting is considered to the other trackers for a fair comparison. Top three results are highlighted in \textcolor{red}{red}, \textcolor{blue}{blue}, and \textcolor{ForestGreen}{green}, and $\ast$ indicates trackers trained exclusively on the GOT-10k training set.\label{MasterTable}}\\

	\hline
	\multirow{2}{*}{\textbf{Trackers}} & 
	\multirow{2}{*}{\textbf{Source}} & \multicolumn{3}{c}{\textbf{GOT-10k $\ast$}} &
	\multicolumn{3}{c}{\textbf{TrackingNet}} & \multicolumn{2}{c}{\textbf{UAV123}}\\ 
	\cmidrule{3-5} \cmidrule{6-8} \cmidrule{9-10}
	& & AO & $\mathrm{SR_{0.5}}$ & $\mathrm{SR_{0.75}}$ & AUC & $\mathrm{P_{n}}$ & P &  AUC & P  \\
	\hline
	\endfirsthead
	
	\multicolumn{10}{c}{Continuation of Table \ref{MasterTable}}\\
	\hline
	\multirow{2}{*}{\textbf{Trackers}} & 
	\multirow{2}{*}{\textbf{Source}} & \multicolumn{3}{c}{\textbf{GOT-10k $\ast$}} &
	\multicolumn{3}{c}{\textbf{TrackingNet}} & \multicolumn{2}{c}{\textbf{UAV123}}\\ 
	\cmidrule{3-5} \cmidrule{6-8} \cmidrule{9-10}
	& & AO & $\mathrm{SR_{0.5}}$ & $\mathrm{SR_{0.75}}$ & AUC & $\mathrm{P_{n}}$ & P &  AUC & P  \\
	\hline
	\endhead
	
	\midrule
	\multicolumn{6}{r}{{Continued on next page...}} \\
	\endfoot
	
	\hline
	\endlastfoot

	\textbf{CPDATrack} & \textbf{Ours} & \color{Red}75.1 & \color{Blue}85.3 &	\color{Red}72.1 & \color{Red}84.1 & \color{Blue}88.9 & \color{ForestGreen}82.8 & 69.4 &  \color{Blue}90.6
	\\
	SIFTrack \cite{KUGARAJEEVAN2025125381} & ESA'25 & \color{ForestGreen}74.6 & \color{Red}85.6 & \color{Blue}71.9 & \color{Red} 84.1 & \color{Red}89.0 & \color{ForestGreen} 82.8 & 68.6 & \color{ForestGreen}90.3 \\
	TBMTrack-256 \cite{ZHANG2025113230} & KBS'25 & 74.4 & 84.5 & 71.7 & 83.6 &
	88.3 &	82.6 & - & -  \\
	ASTrack \cite{CHEN2025126910} & ESA'25 & 73.7 & 84.5 & 70.8 & \color{ForestGreen}83.8  & \color{ForestGreen}88.8 & \color{Blue}83.1 & 69.2 & -\\
	SPformer-256 \cite{CHENG2025128821} & NC'25 & 73.1 & 82.3 & 70.6 & 83.6 & 88.5 & \color{ForestGreen}82.8 & \color{ForestGreen}69.7 & - \\
	PTFTrack \cite{ZHANG2025105431} & IVC'25 & 72.7 & 82.4 & 69.8 & 83.2 & 87.9 & 82.2 & 69.1 & 89.6 \\
	DyTrack-B \cite{Zhu10947615} & IEEE TNNLS'25 & 71.4 & 80.2 & 68.5 & 82.9 & 87.3 & 81.2 & 69.1 & -\\
	CTIFTrack \cite{ZHANG2025125654} & ESA'25 & 71.3 & 81.5 & 66.7 & 82.3 & 87.2 & 80.4 & 67.7 & - \\
	ATrans \cite{LIU2025104553} & VCIR'25 & 71.2 & 	72.3 & 68.6 & 82.7 & 86.8 & 81.4 & - & - \\
	EAPTrack \cite{WU2025104608} & VCIR'25 & 71.0 & 80.9 & 65.6 & 81.5 & 86.5 & 79.5 & 68.1 & 85.4 \\
	MCAT-swin \cite{WANG2025129417} & NC'25 & 70.0 & 80.0 & 63.4 & 81.2 & 85.6 & 79.1 & \color{Red}70.0 & 89.7 \\
	MAPNet-R \cite{SUN2025128785} & NC'25 & 69.5 & 78.4 & 64.9 & 82.3 & 86.4 & 79.6 & 66.3 & 85.7 \\

	AsymTrack-B \cite{Zhu_Tang_Chen_Wang_Wang_Lu_2025} & AAAI'25 & 67.7 & 76.6 & 61.4 & 80.0 & 84.5 & 77.4 & 66.5 & - \\

	GLT \cite{ZHAO2025111} & PRL'25 & 66.5 & 75.8 & 60.2 & 81.8 & 86.4 & 79.9 & - & - \\

	ATFTrans \cite{XuLiang2024} & NCA'24 & 73.6 & 83.2 & \color{Red}72.1 & 83.2 & 87.9 & 82.1 & \color{Blue}69.8 \\

	TiT \cite{YAO_2024_EAAI} & EAAI'24 & 71.6 & 82.5 & 67.4 & 83.4 & 87.8 & 81.8 & 69.3 & \color{Red}91.9\\
	ASAFormer \cite{GONG_2024_KBS} & KBS'24 & 71.3 & 80.0 & 67.6 & 82.2 & 87.3 & 80.6  & 68.6 & 89.9 \\
	CorrFormer \cite{ZHANG_2024_CEE} & CEE'24 & 71.0 & 81.4 & 65.7 & 81.1  & 85.8 & 78.5 & -  & - \\
	
	STTrack \cite{SUN_2024_KBS} & KBS'24 & 70.8 & 80.6 & 66.8 & 82.3 & 86.8 & 80.1 & 69.4 & -  \\
	
	TATrack \cite{HUANG_2024_EAAI} & EAAI'24 & 69.7 & 79.3 & 64.2 & 82.1  & 86.9 & 80.1 & 69.4 &  - \\
	
	UnifiedTT \cite{YU2024104067} & VCIR'24 & 67.3 & 78.8 & 58.1 & - & - & - & 66.2 &  87.3 \\
	
	DeforT \cite{YANG_2024_NN} & NN'24 & - & - & - & 78.6 & 82.7 & 74.6 & 69.1 & 90.2\\
	CRTrack \cite{ZHANG_2024_PR} & PR'24 & 67.0 & 76.8 & 60.3 & 78.0 & 83.0 & 74.2 & 66.9  & 87.4 \\

	SMAT \cite{Gopal_2024_WACV} & WACV'24 & 64.5 & 74.7 & 57.8 & 78.6 & 84.2 & 75.6 & 64.3 & 83.9 \\
	
	SiamSA \cite{YANG2024104237} & VCIR'24 & 54.3 & 65.0 & - & - & - & - & 64.3 & 84.6 \\
	
	SeqTrack-B256 \cite{Chen2023seqtrack} & CVPR'23 & \color{Blue}74.7 & \color{ForestGreen}84.7 & \color{ForestGreen}71.8 & 83.3 & 88.3 & 82.2 & 69.2 & - \\
	GRM \cite{gao2023generalized} & CVPR'23 & 73.4 &	82.9 &	70.4	& \color{Blue}84.0	& 88.7 & \color{Red}83.3  & 69.0 & 89.8  
	\\
	F-BDMTrack-256 \cite{Yang_2023_ICCV} & ICCV'23 & 72.7 & 82.0 & 69.9 & 83.7 & 88.3 & 82.6 & 69.0 & - \\
	TATrack-B \cite{He_Zhang_Xie_Li_Wang_2023} & AAAI'23 & 73.0 & 83.3 & 68.5 & 83.5 & 88.3 & 81.8 & - & - \\
	ROMTrack \cite{Cai_2023_ICCV} & ICCV'23 & 72.9 & 82.9 & 70.2 & 83.6 & 88.4 & 82.7 & - & - \\	
	CTTrack-B \cite{Song_Luo_Yu_Chen_Yang_2023} & AAAI'23 & 71.3 & 80.7 &  70.3 & 82.5 & 87.1 & 80.3 & 68.8 & 89.5
	\\
	MATTrack \cite{Zhao_2023_CVPR} & CVPR'23 & 67.7	 & 78.4	 & -	& 81.9 & 86.8  & - & 68.0 &  - 
	\\
	GdaTFT \cite{Liang_Li_Long_2023} & AAAI'23 & 65.0 &  77.8 & 53.7 & 77.8 & 83.5 & 75.4 & - & - \\			
	BANDT \cite{Yang_2023_IEEETrans} & IEEE TCE'23 & 64.5 & 73.8 & 54.2  & 78.5 & 82.7 & 74.5 & 69.6 & 90.0
	\\
	HiT-Base \cite{Kang_2023_ICCV} & ICCV'23 & 64.0 & 72.1 & 58.1 & 80.0 & 84.4 & 77.3 & 65.6 & - 
	\\
	MixFormer-1k \cite{cui2022mixformer} & CVPR'22 & 71.2 & 79.9 & 65.8 &	82.6 & 87.7 & 81.2 &	68.7 & 89.5
	\\
	OSTrack-256 \cite{ye2022joint} & ECCV'22 & 71.0 &	80.4 &	68.2 & 83.1	& 87.8	& 82.0  & 68.3 & 88.8
	\\
	
	SimTrack-B/16 \cite{chen2022backbone} & ECCV'22 &	69.8 &	78.8 &	66.0  &	82.3 &	86.5 &	80.2 & 	\color{Blue}69.8 &	89.6 \\
	
	DTT \cite{yu2021high} & ICCV'21 & 68.9 & 79.8 & 62.2 & 79.6 & 85.0 & 78.9 & - & -
	\\	
	STARK \cite{yan2021learning} & ICCV'21 & 68.8 &	78.1 &	64.1 &	82.0 &	86.9 &	79.1 &	68.5 & 89.5
	\\	
	TrDiMP \cite{wang2021transformer} & CVPR'21 &	67.1 &	77.7 &	58.3 &	78.4 &	83.3 &	73.1 & 67.5 & -
	\\		
	TransT \cite{chen2021transformer} & CVPR'21 & 67.1 &	76.8 &	60.9  &	81.4 &	86.7 &	80.3  & 68.1 & 87.6
	\\

	\end{longtable}

\end{landscape}

\begin{table}[t!]
	\footnotesize
	\begin{center}
		\caption{Comparison of proposed CPDATrack with state-of-the-art trackers on the LaSOT benchmark using the evaluation metrics $\mathrm{AUC}$ (Area Under the Curve), $\mathrm{P}$ (Precision), and $\mathrm{P_n}$ (Normalized Precision). All results are reported as percentages. Top three results are highlighted in \textcolor{red}{red}, \textcolor{blue}{blue}, and \textcolor{ForestGreen}{green}.}\label{Lasot}%
		\begin{tabular}{l c@{\extracolsep{0.30cm}}l@{\extracolsep{0.30cm}}l@{\extracolsep{0.3cm}}l}
			\midrule
			\multirow{2}{*}{\textbf{Trackers}}  &
			\multirow{2}{*}{\textbf{Source}} &
			\multicolumn{3}{c}{\textbf{LaSOT}} \\ 
			\cmidrule{3-5} 
			& & AUC & $\mathrm{P_{n}}$ & P \\
			\midrule
			\textbf{CPDATrack} & \textbf{Ours} & \color{ForestGreen}69.1 & \color{Red}78.8 & \color{ForestGreen}74.9 \\
			PTFTrack \cite{ZHANG2025105431} & IVC'25 & \color{blue}69.2 & \color{Red}78.8 & \color{Blue}75.1 \\
			CTIFTrack \cite{ZHANG2025125654} & ESA'25 & 66.9 &	77.1 & 72.5 \\
			MAPNet-R \cite{SUN2025128785} & NC'25 & 66.1 & 74.9 & - \\
			EAPTrack \cite{WU2025104608} & VCIR'25 & 66.0 & 75.5 & 70.7 \\
			GLT \cite{ZHAO2025111} & PRL'25 & 65.0 & 74.1 & 68.8 \\
			AsymTrack-B \cite{Zhu_Tang_Chen_Wang_Wang_Lu_2025} & AAAI'25 & 64.7 & 73.0 & 67.8  \\

			GTOT \cite{LI_2024_IS} & IS'24 & 68.2 & - & 74.8 \\
			ASAFormer \cite{GONG_2024_KBS} & KBS'24 &  67.4 & 76.7 & 72.1 \\
			STTrack \cite{SUN_2024_KBS} & KBS'24 & 67.4 & 76.6 & 71.7 \\
			TiT \cite{YAO_2024_EAAI} &  EAAI'24 &  66.7 & 75.4 & 71.6 \\
			TATrack \cite{HUANG_2024_EAAI} & EAAI'24 & 66.1 &  69.4 & 70.3\\
			
			CorrFormer \cite{ZHANG_2024_CEE} & CEE'24 & 64.8  & 74.5 & 68.2 \\
			DeforT \cite{YANG_2024_NN} & NN'24 &  64.3 & 73.0 & 66.3  \\
			CRTrack \cite{ZHANG_2024_PR} & PR'24  & 64.2  & - & 67.1 \\		
			
			UnifiedTT \cite{YU2024104067} & VCIR'24 &  63.8 & 72.7 & 66.6 \\
			SMAT \cite{Gopal_2024_WACV} & WACV'24 & 61.7 & 71.1 & 64.6 \\
			
			TATrack-B \cite{He_Zhang_Xie_Li_Wang_2023} &  AAAI'23 & \color{Red}69.4 & \color{ForestGreen}78.2 & 74.1 
			\\
			CTTrack-B \cite{Song_Luo_Yu_Chen_Yang_2023} &  AAAI'23 & 67.8 & 77.8 & 74.0 
			\\
			MATTrack \cite{Zhao_2023_CVPR} & CVPR'23 & 67.8 & 77.3 &  - 
			\\

			HiT-Base \cite{Kang_2023_ICCV} & ICCV'23 & 64.6 & 73.3 & 68.1 \\
			BANDT \cite{Yang_2023_IEEETrans} & IEEE TCE'23 &  64.4 & 72.4 & 67.0
			\\
			GdaTFT \cite{Liang_Li_Long_2023} & AAAI'23 &  64.3 & 68.0 & 68.7 \\

			MixFormer-22k \cite{cui2022mixformer} & CVPR'22 & \color{Blue}69.2 &	\color{Blue}78.7 &	74.7 	
			\\
			OSTrack-256 \cite{ye2022joint} & ECCV'22 & \color{ForestGreen}69.1 & \color{Blue}78.7 & \color{Red}75.2 \\			
			CSWinTT \cite{song2022transformer} & CVPR'22 &	66.2 &	75.2 &	70.9 \\			
			SparseTT \cite{fu2022sparsett} & IJCAI'22 & 	66.0 &	74.8 &	70.1 \\
			
			STARK \cite{yan2021learning} & ICCV'21 & 	67.1 &	77.0 &	72.2
			\\			
			TransT \cite{chen2021transformer} & CVPR'21 & 	64.9 &	73.8 &	69.0
			\\
			TrDiMP \cite{wang2021transformer} & CVPR'21 &	63.9 &	73.0 &	61.4
			\\

			\bottomrule
		\end{tabular}
	\end{center}
	
\end{table}

\subsection{Ablation Studies}\label{sec4subsec3}
Extensive ablation experiments were conducted by training the models on the GOT-10k training set and evaluating their tracking effectiveness on the corresponding test set, with the objective of validating the tracker’s design and the selected parameter configurations.

\begin{table}[t]
	\footnotesize
	\begin{center}
		\caption{Ablation study on the contributions of the Context-Aware Token Pruning (CATP) module and the Discriminative Selective Attention (DSA) mechanism, evaluated with and without the Spatial Confidence Zone (SCZ). Relative improvements over the baseline are indicated.}\label{tabel:Ablation1}%
		\begin{tabular}{cclll}
			\toprule
			\multirow{2}{*}{\textbf{Model}}&
			\multirow{2}{*}{\textbf{Modules}}& \multicolumn{3}{c}{\textbf{GOT-10k}}\\ 
			\cmidrule{3-5}
			& &  AO & $\mathrm{SR_{0.5}}$ & $\mathrm{SR_{0.75}}$ \\ 
			\midrule
			A& Baseline & 73.0 &	83.3 &	70.3 \\
			B & + CATP &  73.6 \textcolor{blue}{\scriptsize ↑0.6} & 83.6 \textcolor{blue}{\scriptsize ↑0.3} & 71.5 \textcolor{blue}{\scriptsize ↑1.2}\\
			C & + CATP + DSA (w/o SCZ) &  74.3 \textcolor{blue}{\scriptsize ↑1.3}&	84.4 \textcolor{blue}{\scriptsize ↑0.8} &	72.2 \textcolor{blue}{\scriptsize ↑1.9}\\
			D & + CATP + DSA (w/ SCZ) &  \textbf{75.1} \textcolor{blue}{\scriptsize ↑2.1} &	\textbf{85.3} \textcolor{blue}{\scriptsize ↑2.6} &	\textbf{72.1} \textcolor{blue}{\scriptsize ↑1.8}\\
			\midrule
		\end{tabular}
	\end{center}
\end{table}

\subsubsection{Contributions of Individual Components }\label{sec4subsec3subsec1}
The CATP module and the DSA mechanism constitute the core components of the proposed CPDATrack. Furthermore, the tracker’s performance is enhanced by mitigating the influence of distractors through the introduction of a SCZ, centered around the token with the highest aggregated target probability. To evaluate the contribution of these components, module-wise ablation experiments were conducted, and the results are summarized in Table \ref{tabel:Ablation1}.

The baseline tracker, denoted as Model A, processes search, initial, and dynamic template tokens without employing any token pruning mechanism or techniques to mitigate interference from background tokens. In Model B, the integration of the CATP module reduces background interference by pruning less informative tokens, thereby improving overall tracking performance. In Model C, the interference of background tokens is further minimized through the DSA module, enhancing the tracker's discriminative capability and leading to better performance. Finally, Model D (proposed tracker) achieves additional performance gains by mitigating the impact of distractors through the introduction of the SCZ in conjunction with the DSA module. This ablation experiments clearly indicates the contribution of each modules of the proposed tracker. 

\begin{table}[t]
	\footnotesize
	\begin{center}
		\caption{Ablation study on the effect of the Context-Aware Token Pruning (CATP) module in the proposed CPDATrack.}\label{tabel:Ablation2}%
		\begin{tabular}{clll}
			\midrule
			\multirow{2}{*}{\textbf{Token Pruning}} &  \multicolumn{3}{c}{\textbf{GOT-10k}}\\ 
			\cmidrule{2-4}
			&  AO & $\mathrm{SR_{0.5}}$ & $\mathrm{SR_{0.75}}$ \\
			\midrule

			CPDATrack without pruning & 73.8 &	84.2 &	71.9 \\
			CPDATrack with conventional pruning & 73.7 & 83.5 & 71.0 \\
			CPDATrack with CATP & \textbf{75.1} &	\textbf{85.3} &	\textbf{72.1} \\

			\midrule
		\end{tabular}
	\end{center}
\end{table}

\subsubsection{Effectiveness of Context-Aware Token Pruning}\label{sec4subsec3subsec2}

\begin{figure}[t]%
	\centering
	\includegraphics[width=0.48\textwidth]{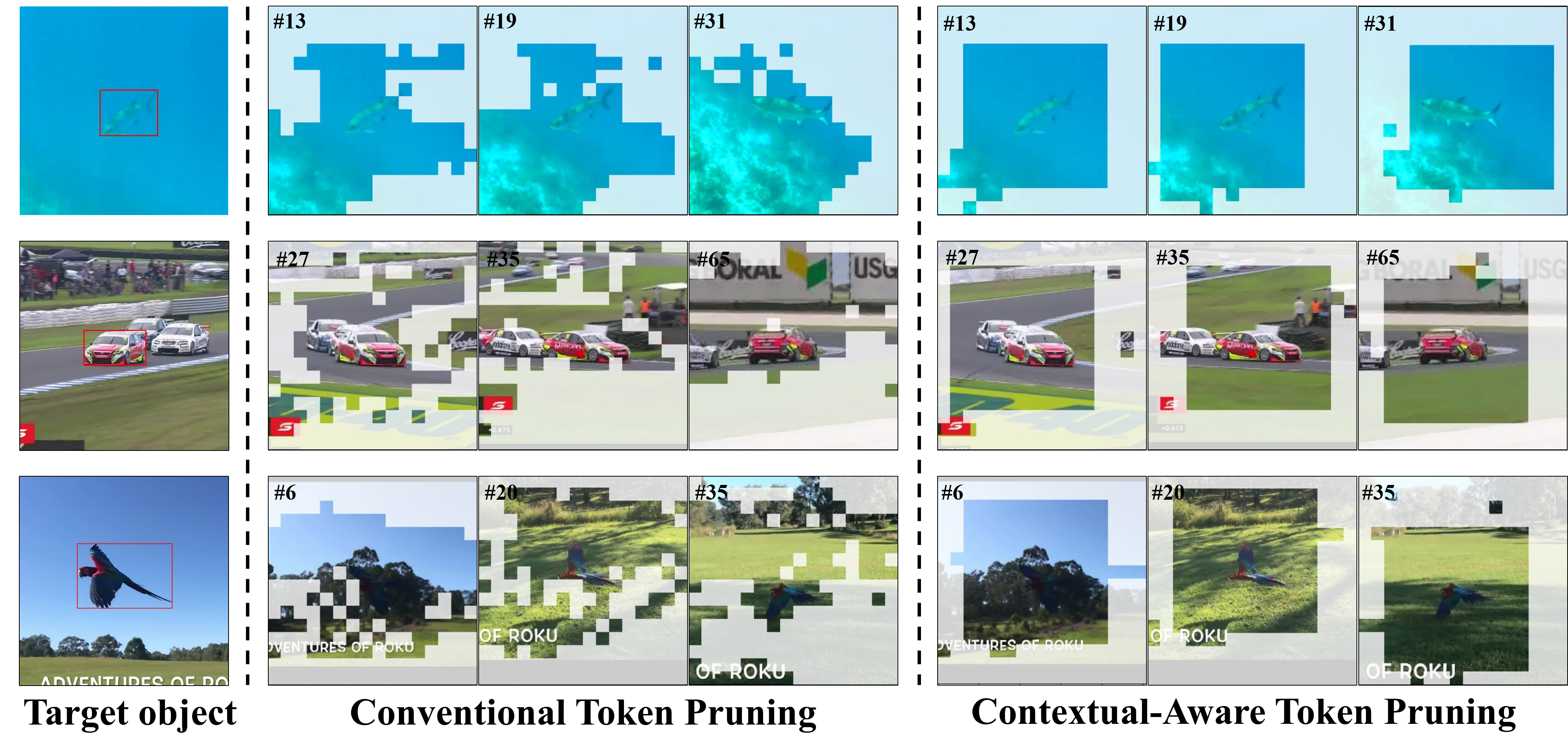}
	\caption{Qualitative comparison of conventional and proposed context-aware pruning. Conventional pruning indiscriminately removes tokens, including valuable low-attention ones and those near the target, leading to contextual information loss, whereas context-aware pruning selectively retains tokens within the contextual zone, preserving target features and surrounding context.}
	\label{fig:pruning}
\end{figure}

Previous token pruning techniques, referred to as conventional pruning, remove less informative background tokens from the search region without considering the potential loss of important contextual information surrounding the target. Although low-attention patches may appear individually weak, they still contribute to the overall semantic representation of the target. Likewise, nearby background tokens provide essential spatial context that aids in distinguishing the target from its surroundings. The proposed CATP module addresses this limitation by removing less informative background tokens while preserving a contextual zone centered around the token with the highest aggregated target probability. To validate the effectiveness of the proposed CATP module, ablation experiments were conducted, and quantitative results presented in Table \ref{tabel:Ablation2} and qualitative comparisons shown in Fig. \ref{fig:pruning}.

\begin{table}[b]
	\footnotesize
	\begin{center}
		\caption{Ablation study on the effect of Contextual Zone (CZ) window size during pruning.}\label{tabel:Ablation3}%
		\begin{tabular}{cclll}
			\midrule
			\multirow{2}{*}{\textbf{CZ Size}} & \multicolumn{3}{c}{\textbf{GOT-10k}}\\ 
			\cmidrule{2-4}
			&   AO & $\mathrm{SR_{0.5}}$ & $\mathrm{SR_{0.75}}$ \\
			\midrule	
			$ 7 \times 7$ &  73.1 & 83.0 & 70.7\\
			$ 9 \times 9$ &  74.6 &	85.0 &	71.1\\
			$ 11 \times 11$ & \textbf{75.1} &	\textbf{85.3} &	\textbf{72.1} \\
			$ 13 \times 13$ &  72.9 &	82.9 & 70.6 \\
			
			\midrule
		\end{tabular}
	\end{center}
\end{table}

\begin{table*}[t]
	\scriptsize
	\begin{center}
		\caption{Ablation study on the Discriminative Selective Attention (DSA) mechanism. Model performance is compared based on different search-to-template attention blocking and allowing configurations.}\label{tabel:Ablation4}
		\begin{tabular}{cccccccc}
			\midrule
			\textbf{Model} & \textbf{First to $l^{th}$  layer} & \textbf{  $(l+1)^{th}$ to Last layer} & \textbf{CATP} &\textbf{SCZ} & \multicolumn{3}{c}{\textbf{GOT-10k}} \\
			\cmidrule{6-8}
			& & & & & AO & SR$_{0.5}$ & SR$_{0.75}$ \\
			\midrule
			1 & Allowed & Allowed & - & - &  73.0 & 83.3 & 70.3  \\
			2 & Blocked & Blocked & - & - &  72.7 & 82.8 & 69.8  \\
			3 & Blocked & Only target tokens allowed & \checkmark &  - &  74.3 & 84.4 & \textbf{72.2 } \\
			4 & Blocked & Only target tokens allowed & \checkmark & \checkmark &  \textbf{75.1} & \textbf{85.3} & 72.1  \\
			5 & Allowed & Only target tokens allowed & \checkmark & \checkmark &  73.5 & 84.0 & 70.5  \\
			
			\bottomrule
		\end{tabular}
	\end{center}
\end{table*}

Based on the experimental results presented in Table \ref{tabel:Ablation2}, it is evident that applying conventional pruning techniques to the proposed CPDATrack, without altering any other settings, leads to a decline in performance. The removal of important contextual cues during pruning results in the tracker struggling to accurately identify the target. The superior performance of CPDATrack with the CATP module clearly underscores the significance of preserving contextual cues in tracking. The qualitative comparison in Fig. \ref{fig:pruning} further illustrates that conventional pruning removes background tokens surrounding the target, even when they contain useful contextual cues. It can also be observed that some low-confidence target tokens are pruned in conventional methods, whereas such cases do not occur in the proposed tracker.

In the proposed CATP module, tokens within the contextual zone are not pruned. An additional ablation experiment was conducted to analyse the tracker’s performance with respect to different sizes of the contextual zone during pruning. As summarized in Table \ref{tabel:Ablation3}, the tracker achieves the best results with an $11 \times 11$ contextual zone, which effectively captures sufficient spatial context around the target while avoiding excessive background noise. When the contextual zone is smaller (e.g., $7 \times 7$), it may prune important surrounding tokens, leading to a loss of valuable contextual information and weakening the tracker’s ability to distinguish the target from the background. Conversely, a larger zone (e.g., $13 \times 13$) incorporates a significant amount of irrelevant background tokens, introducing noise that can distract the tracker and reduce its discriminative capability. Thus, the $11 \times 11$ zone provides a balanced trade-off, retaining critical contextual information while minimizing interference from irrelevant regions, resulting in more accurate and robust tracking.

\subsubsection{Effectiveness of Discriminative Selective Attention}\label{sec4subsec3subsec3}

\begin{figure}[t!]%
	\centering
	\includegraphics[width=0.48\textwidth]{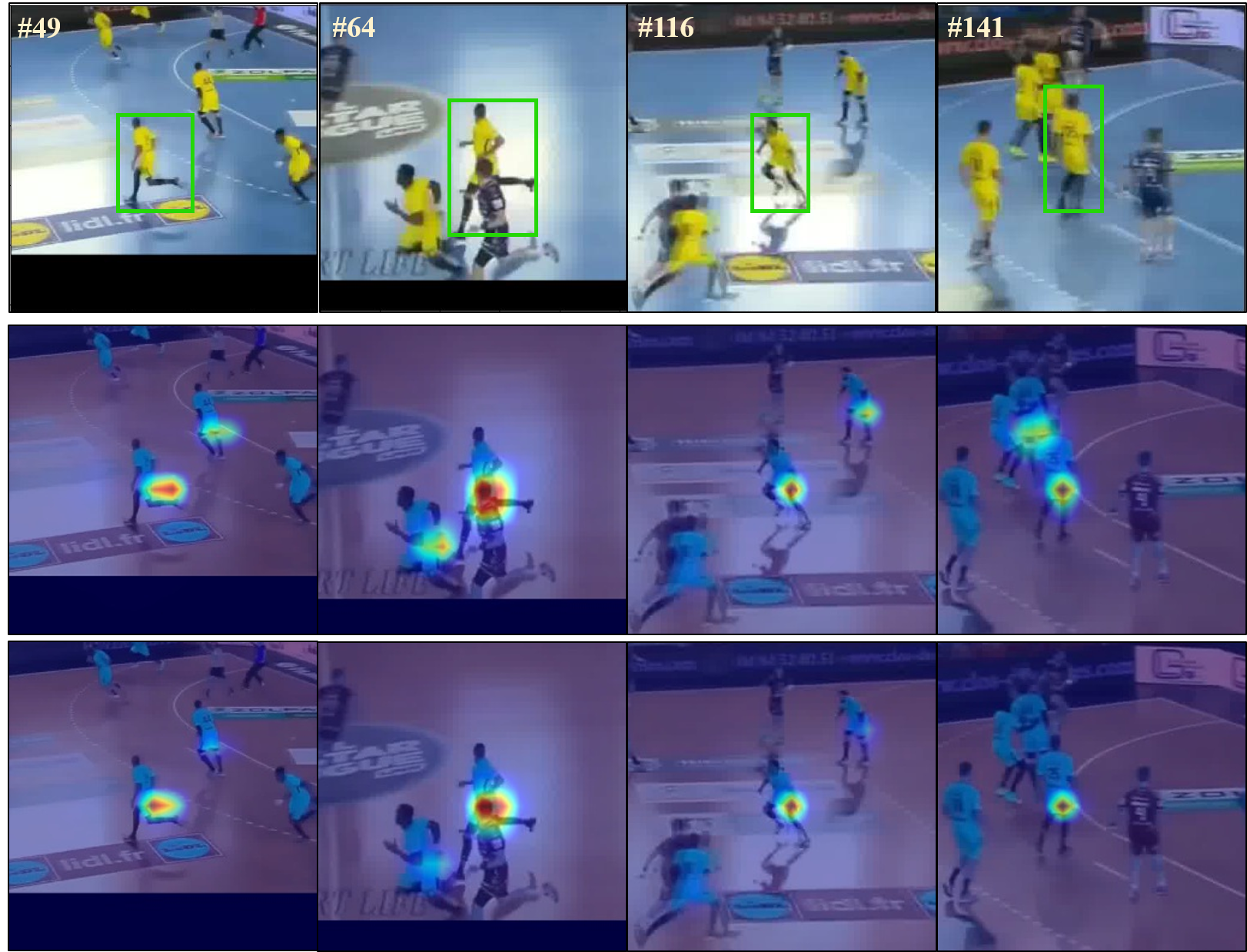}
	\caption{Qualitative comparison of classification heatmaps without (middle row) and with (bottom row) the Spatial Confidence Zone (SCZ). Without the SCZ, the tracker incorrectly identifies the distractor as part of the target, as distractor tokens weaken the target–template feature representation. In contrast, the proposed tracker restricts attention to tokens within the SCZ, effectively suppressing distractor influence.}
	\label{fig:SAM_heatmap}
\end{figure}

The proposed DSA mechanism further reduces the impact of background tokens and minimises the influence of distractor objects. To validate the effectiveness of the DSA mechanism, an ablation study was conducted by selectively blocking or allowing attention between different groups of tokens, and the results are presented in Table \ref{tabel:Ablation4}.

In these ablation experiments, the tracker’s performance is compared under various search-to-template attention blocking and allowing configurations. The baseline model, denoted as Model \#1 in Table \ref{tabel:Ablation4}, allows all search region tokens to attend to both the initial template and dynamic target template across all encoder layers. In Model \#2, the attention from all search tokens to both the initial target and dynamic target templates is completely blocked across all layers to mitigate the influence of background tokens. However, Model \#2 exhibited inferior performance compared to Model \#1, as the search region contains valuable target tokens that are crucial for learning the target’s appearance variations, which are more significant than suppressing background interference. In Model \#3, the search-to-template attention is blocked from the first layer up to layer $l$. Additionally, a considerable number of background tokens are removed by the CATP module, while target tokens within the search region are identified using the TPE module. In the subsequent layers, only the identified target search tokens are allowed to attend to both the initial and dynamic template tokens. Since Model \#3 reduces background interference while enabling the tracker to effectively capture target features, it showed better performance. The influence of distractors in Model \#3 is mitigated in Model \#4 by selecting target search tokens within a narrow SCZ, centered on the token with the highest aggregated target probability. To validate the importance of blocking search-to-template attention in the early layers, Model \#5 is implemented by allowing early-layer attention, which reduces performance compared to Model \#4 and confirms that completely blocking attention in the early encoder layers is crucial, as early feature representations lack sufficient discriminative strength and are prone to contamination from background tokens.

\begin{table}[t]
	\footnotesize
	\begin{center}
		\caption{Ablation study on the effect of the Spatial Confidence Zone (SCZ) size.}\label{tabel:Ablation5}%
		\begin{tabular}{clll}
			\midrule
			\multirow{2}{*}{\textbf{SCZ Size}} &  \multicolumn{3}{c}{\textbf{GOT-10k}}\\ 
			\cmidrule{2-4}
			&  AO & $\mathrm{SR_{0.5}}$ & $\mathrm{SR_{0.75}}$ \\
			\midrule	
			
			$ 5 \times 5$ & 73.5 & 83.7 & 71.3\\
			$ 7 \times 7$ & \textbf{75.1} &	\textbf{85.3} &	\textbf{72.1} \\
			$ 9 \times 9$ & 73.5 & 83.6 & 71.5 \\
			
			\midrule
		\end{tabular}
	\end{center}
\end{table}

To demonstrate that the proposed tracker effectively mitigates the influence of distractors, the classification heatmaps of a tracking sequence with and without the SCZ are presented in Fig. \ref{fig:SAM_heatmap}.  As shown in the middle row of Fig. \ref{fig:SAM_heatmap}, when all target search tokens are allowed to attend to the template tokens, distractors tend to weaken the target-specific features. In contrast, when the target search tokens are restricted to the narrow SCZ, the influence of distractors is significantly reduced.

Another ablation experiment is conducted to investigate the effect of the SCZ size on the tracker’s performance.  As summarized in Table \ref{tabel:Ablation5}, the tracker achieves the best results with a $7 \times 7$ SCZ, which effectively retains relevant target tokens while suppressing distant distractors. When the SCZ is too small (e.g., $5 \times 5$), it may exclude essential target tokens, leading to degraded performance. Conversely, a larger SCZ (e.g., $9 \times 9$) tends to include more distractor tokens, introducing noise that weakens the tracker’s discriminative capability. Therefore, a $7 \times 7$ SCZ provides an optimal balance, preserving critical target information while reducing interference from irrelevant distractor tokens, resulting in more accurate and robust tracking.

\subsubsection{Optimal Placement of the Target Probability Estimation Module}\label{sec4subsec3subsec4}

\begin{table}[t]
	\footnotesize
	\begin{center}
		\caption{Ablation study on the placement of the Target Probability Estimation (TPE) module.}
		\label{tabel:Ablation6}%
		\begin{tabular}{clll}
			\midrule
			\multirow{2}{*}{\textbf{Layers between}} &  \multicolumn{3}{c}{\textbf{GOT-10k}}\\
			\cmidrule{2-4}
			&  AO & $\mathrm{SR_{0.5}}$ & $\mathrm{SR_{0.75}}$ \\
			\midrule
			
			3 and 4 & 74.2 & 84.5 & 71.5\\
			4 and 5 & \textbf{75.1} & \textbf{85.3} & 72.1 \\
			5 and 6 & 74.5 & 84.7 & \textbf{72.4} \\
			6 and 7 & 73.7 & 84.0 & 70.7 \\
			7 and 8 & 73.0 & 82.8 & 70.3 \\

			\midrule
		\end{tabular}
	\end{center}
\end{table}

The Target Probability Estimation (TPE) module plays an important role in the proposed tracker, as it guides both the CATP module and the DSA mechanism. The TPE module estimates the target likelihood probability of each token in the search region based on the features from the preceding layer where it is integrated. It is positioned between two specific encoder layers, and an ablation study is conducted to justify this placement as optimal.

Based on the experimental results presented in Table \ref{tabel:Ablation6}, the proposed tracker exhibits reduced performance when the TPE module is placed between the 3rd and 4th encoder layers, as the semantic information captured by the 3rd layer is insufficient to accurately estimate the target probability. Placing the TPE module between the 4th and 5th layers yields the best performance, achieving a balance between accurate token partitioning and effective propagation of target-aware features in the subsequent layers. However, when the TPE module is placed after the middle layers (beyond the 6th layer), the tracker’s performance begins to decrease slightly. One possible reason is that token pruning is also performed at the position where the TPE module is placed. As a large number of tokens are removed in a single shot, the subsequent layers have limited capacity to refine and propagate the partitioned token information, resulting in a decline in overall performance.

\subsection{Comparison of Computational Complexity}

\begin{table}[!t]
	\footnotesize
	\begin{center}
		\caption{Efficiency comparison of the proposed tracker with existing token pruning and background interference reduction methods, reporting the number of parameters, FLOPs, tracking speed (FPS), and Average Overlap (AO) on the GOT-10k benchmark.}\label{Efficiency}%
		\begin{tabular}{l l l l l}
			\midrule
			\multirow{2}{*}{\textbf{Tracker}} & \textbf{Parameters} & \textbf{FLOPs}& \multirow{2}{*}{\textbf{FPS}}  &  \multirow{2}{*}{\textbf{AO}}\\
			&   \textbf{(M)} & \textbf{(G)} &  & \\
			\midrule
			
			CPDATrack (ours) & 94.66 & 27.80 & 43 & 75.1 \\
			SIFTrack \cite{KUGARAJEEVAN2025125381} & 92.63 & 33.82 & 20 & 74.6 \\
			
			GRM \cite{gao2023generalized} & 99.83 & 30.90 & 27 & 73.4 \\
			OSTrack-256 \cite{ye2022joint} & 92.83  & 21.50  &  50 & 71.0\\

			\midrule
		\end{tabular}
	\end{center}
\end{table}

The computational complexity of the proposed CPDATrack is compared with existing token pruning and background interference reduction methods. All comparisons are conducted using the same search region size ($256 \times 256$). Although several similar approaches exist, they were not considered in this comparison as their models are not publicly available. 

As summarized in Table~\ref{Efficiency}, proposed CPDATrack achieves the highest AO of 75.1\% on GOT-10k while requiring only 27.8 GFLOPs and running at 43 FPS, demonstrating both superior accuracy and efficiency. OSTrack also employs a token pruning mechanism and achieves higher efficiency than our tracker, as it does not use a dynamic template and therefore processes 64 fewer tokens. GRM does not employ any pruning mechanism but includes a background interference reduction technique. In contrast, SIFTrack incorporates both pruning and background interference reduction mechanisms. Compared with these trackers, the proposed tracker achieves superior tracking performance while maintaining a reasonable computational cost, making CPDATrack well-suited for real-time applications.

\begin{figure*}[t!]%
	\centering
	\includegraphics[width=0.98\textwidth]{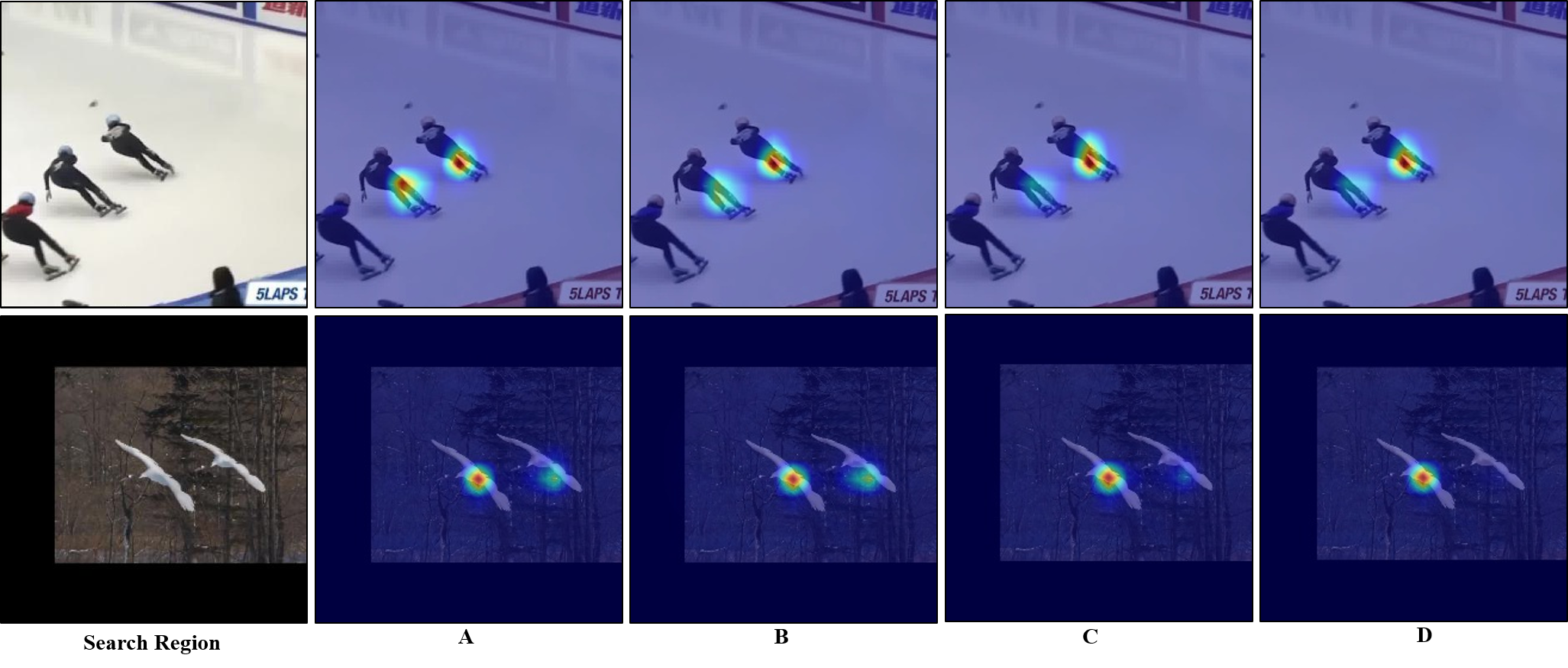}
	\caption{Visualization demonstrating the effectiveness of CPDATrack through comparative classification heatmaps across different model configurations. Model A : attends to all search tokens; Model B: applies context-aware token pruning (CATP); Model C: incorporates selective attention mechanism (SAM); Model D: applies SAM with a Spatial Confidence Zone (SCZ) to focus on target-relevant regions and suppress distant distractors.}
	\label{fig:models_heatmap}
\end{figure*}

\begin{figure}[t!]%
	\centering
	\includegraphics[width=0.48\textwidth]{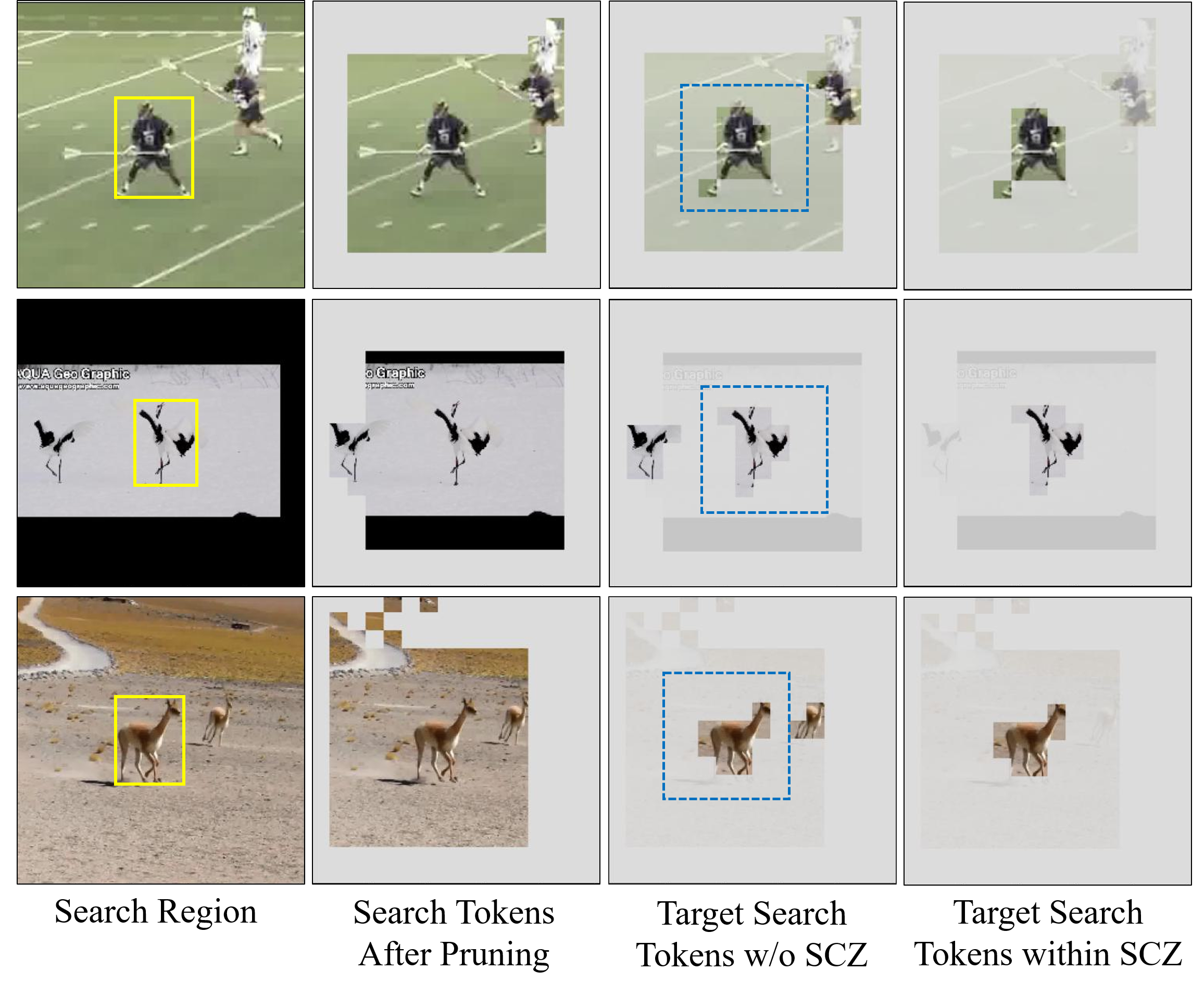}
	\caption{Visualization illustrating the effectiveness of context-aware token pruning, target token identification, and the selection of target tokens within the SCZ region (indicated by the blue box) to avoid distractors. }
	\label{fig:selected_tokens}
\end{figure}

\subsection{Discussion}

In this work, a novel one-stream Transformer tracking model is proposed to mitigate background token interference, suppress distractor effects, and enhance computational efficiency.  To validate the effectiveness of the proposed methodology, the classification heatmaps of the proposed tracker (denoted as Model D) and other configurations are presented in Fig. \ref{fig:models_heatmap}. In Model A, which attends to all search tokens for both the initial and dynamic templates across all layers, background and distractor tokens interfere with the target feature representation, causing the tracker to fail in distinguishing the actual target from distractors. In Model B, the CATP module removes background tokens while preserving contextual cues around the target, thereby enhancing the tracker’s discriminative capability. However, distractors are still occasionally identified as the target object, albeit with slightly lower confidence. In Model C, background interference is further reduced by allowing only target-relevant search tokens to attend to the template tokens. Finally, in Model D, it is evident that distractor interference is almost completely eliminated by applying DSA with SCZ, which restricts attention to search tokens within a spatial confidence region around the target and prevents distant distractors from contaminating the features.

To illustrate the effectiveness of token pruning and target token selection within the search region, Fig. \ref{fig:selected_tokens} presents the visualization of retained tokens after pruning, the target tokens identified by the TPE module, and those selected based on the SCZ. Based on the figure, it can be clearly observed that the pruned tokens correspond to less-informative background regions, demonstrating the effectiveness of context-aware pruning in preserving the target and its contextual information. The identified target tokens within the search region demonstrate the effectiveness of the TPE module. However, since similar distractor tokens are also identified as targets, selecting target tokens from the narrow SCZ region demonstrates how the proposed tracker effectively avoids distractor interference.

Overall, CPDATrack achieves the highest accuracy, confirming that context-aware pruning combined with discriminative selective attention enhances target-focused feature extraction and improves tracking robustness. CPDATrack not only improves tracking accuracy but also demonstrates high computational efficiency compared with similar state-of-the-art trackers, requiring fewer FLOPs while maintaining a real-time tracking speed of 43 FPS.

\begin{figure}[t]%
	\centering
	\includegraphics[width=0.48\textwidth]{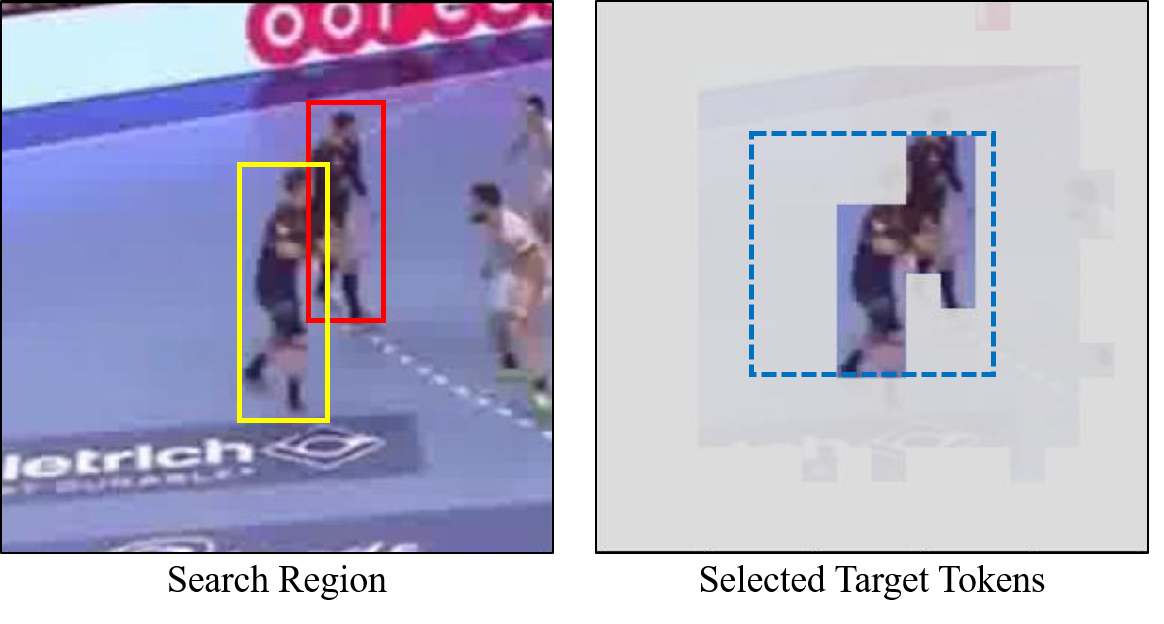}
	\caption{Illustration of a challenging scenario in CPDATrack. When a similar distractor (yellow box) is adjacent to the target (red box), the model may incorrectly select the distractor as a target, causing tracking interference and drift.}
	\label{fig:Fail_case}
\end{figure}

Although the proposed CPDATrack demonstrates superior performance by effectively suppressing background and distractor interference, its long-term tracking performance remains modest compared to other recent trackers. This limitation arises when a similar distractor appears in close proximity to the target, as shown in Fig. \ref{fig:Fail_case}. In such cases, the proposed tracker fails to distinguish the distractor from the target because both lie within the narrow SCZ region, leading to interference and causing the tracker to drift toward the distractor, thereby reducing overall tracking performance. As future work, we plan to incorporate more robust distractor modeling to enhance the tracker’s ability to distinguish the target from nearby distractors, thereby ensuring more consistent and accurate performance over long sequences. Another possible reason is that the TPE module identifies target search tokens based on the features of the initial and dynamic templates, and an incorrect selection of the dynamic template during long tracking sequences can lead to tracking failure.

\section{Conclusion}\label{sec5}
In this study, a novel one-stream Transformer-based tracking approach, termed CPDATrack, is proposed to mitigate the adverse effects of background and distractor interference while enhancing computational efficiency. The proposed tracker prunes less informative background tokens in an intermediate layer while preserving contextual cues surrounding the target. In addition, the discriminative selective attention mechanism further  suppresses background token interference. Moreover, the influence of distractors is reduced by selecting the target search tokens from a narrowly defined spatial confidence zone and allowing them to attend only to the initial and dynamic template tokens. Evaluation on four benchmark datasets demonstrates that CPDATrack achieves superior tracking performance at 43 FPS, with particularly strong results on GOT-10k where it attains an average overlap of 75.1\%. These findings underscore the effectiveness of context-aware pruning and discriminative selective attention in enabling precise and reliable target tracking.

\section*{CRediT authorship contribution statement}
\textbf{Janani Kugarajeevan}: Conceptualization, Methodology,  Investigation, Implementation, Validation, Writing - Original Draft, Visualization. \textbf{Kokul Thanikasalam}: Conceptualization, Investigation, Methodology, Supervision, Writing - Original Draft. \textbf{Amirthalingam Ramanan}: Conceptualization, Supervision, Writing - Review \& Editing. \textbf{Subha Fernando}: Conceptualization, Supervision, Writing - Review \& Editing

\section*{Declaration of competing interest}
The authors declare that they have no known competing financial interests or personal relationships that could have appeared to influence the work reported in this paper.

\section*{Data availability}
Following publicly available benchmark datasets are used in this study.
\begin{itemize}
	\itemsep0em
	\item GOT-10k: \url{http://got-10k.aitestunion.com}
	\item TrackingNet: \url{https://tracking-net.org}
	\item LaSOT: \url{http://vision.cs.stonybrook.edu/~lasot}
	\item UAV123: \url{https://cemse.kaust.edu.sa/ivul/uav123}
\end{itemize}
The code for this work is publicly available at \url{https://github.com/JananiKugaa/CPDATrack.git}

\bibliographystyle{elsarticle-num} 
\bibliography{bibliography}



\end{document}